%% file: main.tex
\documentclass[10pt,twocolumn,letterpaper]{article}

\usepackage{cvpr}
\usepackage{times}
\usepackage{epsfig}
\usepackage{graphicx}
\usepackage{amsmath}
\usepackage{amssymb}
\usepackage{mathtools}
\usepackage{breqn}

\usepackage{url}
\usepackage{mathrsfs}
\usepackage[table]{xcolor}
\usepackage{dsfont}
\usepackage{xspace}
\usepackage{listings}
\usepackage{numprint}
\usepackage{multirow}
\usepackage{setspace}
\usepackage{lipsum}
\usepackage{caption}
\usepackage{booktabs}
\usepackage{microtype}
\usepackage{graphicx}
\usepackage{sidecap}
\usepackage{subfigure}
\usepackage{balance}
\usepackage{pifont}
\usepackage{caption}

\usepackage{color}
\definecolor{forestgreen}{rgb}{0.0, 0.27, 0.13}
\definecolor{brickred}{rgb}{0.8, 0.25, 0.33}
 
\usepackage[pagebackref=false,breaklinks=true,letterpaper=true,colorlinks,bookmarks=false]{hyperref}

\cvprfinalcopy % *** Uncomment this line for the final submission

\def\cvprPaperID{2218} % *** Enter the CVPR Paper ID here

\DeclareMathOperator*{\argmax}{arg\,max}

\newcommand{\mcM}{\mathcal{M}}  % movie
\newcommand{\mcT}{\mathcal{T}}  % trimmed clips set
\newcommand{\mcC}{\mathcal{C}}  % characters
\newcommand{\mcP}{\mathcal{P}}  % character pairs
\newcommand{\mcA}{\mathcal{A}}  % interaction labels
\newcommand{\mcR}{\mathcal{R}}  % relationship labels
\newcommand{\mcO}{\mathcal{O}}  % overlapping labels
\newcommand{\bW}{\mathbf{W}}
\newcommand{\bw}{\mathbf{w}}
\newcommand{\bb}{\mathbf{b}}
\newcommand{\real}{\mathbb{R}}

\newcommand{\cmark}{\ding{51}}

\ifcvprfinal\fi
\title{Learning Interactions and Relationships between Movie Characters}
\author{
Anna Kukleva$^{1,2}$
\hspace{1cm}
Makarand Tapaswi$^1$%
\hspace{1cm}
Ivan Laptev$^1$\\
$^1$Inria Paris, France\hspace{0.5cm}
$^2$Max-Planck-Institute for Informatics, Saarbr\"ucken, Germany\\
{\tt\small akukleva@mpi-inf.mpg.de, \{makarand.tapaswi,ivan.laptev\}@inria.fr}
}

\begin{document}

% \maketitle
%\thispagestyle{empty}

\twocolumn[{%
\renewcommand\twocolumn[1][]{#1}%
\maketitle

\vspace{-0.2cm}
\centering
\includegraphics[width=\linewidth]{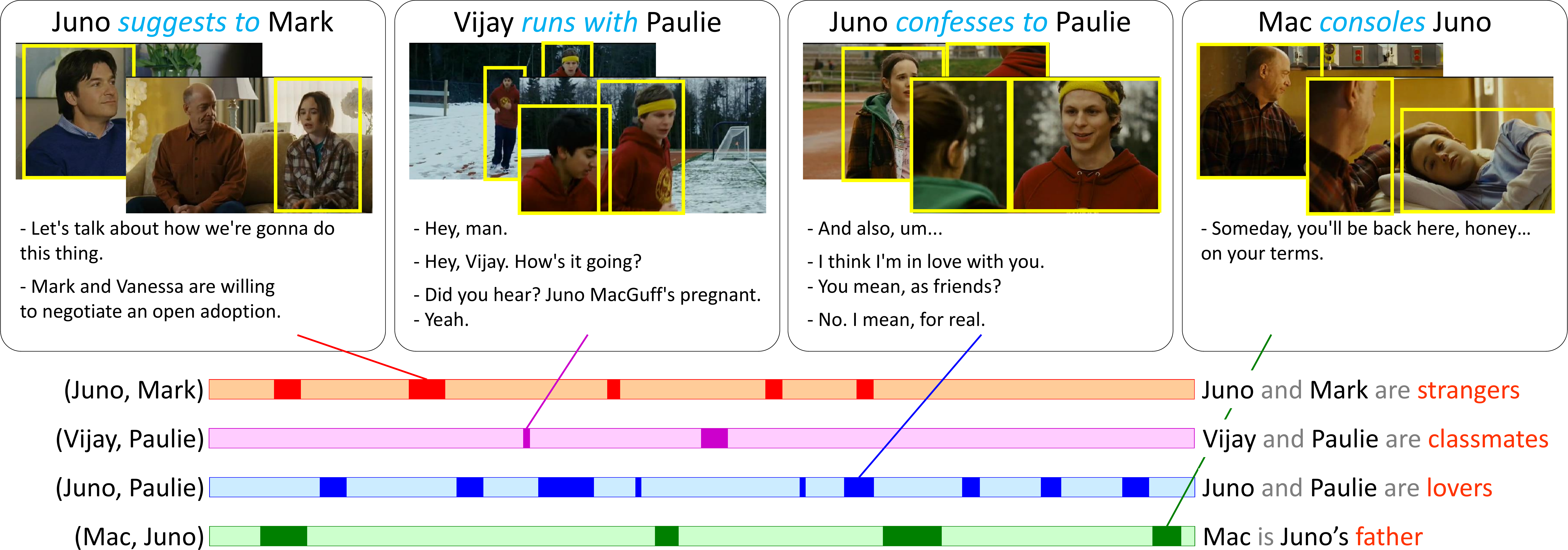}
\vspace{-0.4cm}
\captionof{figure}{
The goal of this work is to jointly predict interactions and relationships between all  characters in movies.
Some interactions are based on dialog (\eg~suggests, confesses), some are primarily visual (\eg~runs with), and others are based on a fusion of both modalities (\eg~consoles).
The colored rows at the bottom highlight when a pair of characters appear in the movie timeline.
Their (directed) relationships are presented at the right.
Example clips obtained from the movie \emph{Juno, 2007}.
}
\label{fig:intro}
% \vspace{0.8cm}
\vspace{0.4cm}
}]

%%%%%%%%% ABSTRACT
\begin{abstract}
\vspace{-2mm}
Interactions between people are often governed by their relationships.
On the flip side, social relationships are built upon several interactions.
Two strangers are more likely to greet and introduce themselves while becoming friends over time.
We are fascinated by this interplay between interactions and relationships, and believe that it is an important aspect of understanding social situations.
In this work, we propose neural models to learn and jointly predict interactions, relationships, and the pair of characters that are involved.
We note that interactions are informed by a mixture of visual and dialog cues, and present a multimodal architecture to extract meaningful information from them.
Localizing the pair of interacting characters in video is a time-consuming process, instead, we train our model to learn from clip-level \emph{weak} labels.
We evaluate our models on the MovieGraphs dataset and show the impact of modalities, use of longer temporal context for predicting relationships, and achieve encouraging performance using weak labels as compared with ground-truth labels.
Code is online.%
\footnote{\url{https://annusha.github.io/LIReC/}}
% \vspace{-10mm}
\end{abstract}

%%%%%%%%% BODY TEXT
\vspace{-5mm}
\input{intro}

\input{relwork}

\input{model}

\input{eval}

\input{conc}

% \balance
{\small
\bibliographystyle{ieee_fullname}
\bibliography{refs}
}

\newpage
{\normalsize
\appendix
\input{appendix.tex}

}
\end{document}

%% file: intro.tex
\section{Introduction}
\label{sec:intro}
% \footnotetext[1]{work done during an internship at Inria.}
% -- interactions and relationships
A salient aspect of being human is our ability to interact with other people and develop various relationships over the period of our lives.
While some relationships drive the typical interactions experienced by a pair of people in a top-down manner (\eg~parents customarily love and nurture their children); almost all social (non-family) relationships are driven through bottom-up interactions (\eg~strangers become friends over a good chat or a shared drink)~\cite{hinde1976}.
For an intelligent agent to truly be a part of our lives, we will need it to assimilate this complex interplay and learn to behave appropriately in different social situations.

% -- movie understanding about people
We hypothesize that a first step in this direction involves learning how people interact and what their relationships might be.
However, training machines with live, real world, experience-based data is an extremely complicated proposition.
Instead, we rely on movies that provide snapshots into key moments of our lives, portraying human behavior at its best and worst in various social situations~\cite{vicol2018moviegraphs}.

Interactions and relationships have been addressed separately in literature.
Interactions are often modeled as simple actions~\cite{gu2018ava,alonso2010high_five}, and relationships are primarily studied in still images~\cite{li2017dual,sun2017domain} and recently in videos~\cite{liu2019social}.
However, we believe that a complete understanding of social situations can only be achieved by modeling them jointly.
For example, consider the evolution of interactions and relationships between a pair of individuals in a romantic movie.
We see that the characters first meet and talk with each other and gradually fall in love, changing their relationship from strangers to friends to lovers.
This often leads to them getting married, followed subsequently by arguments or infidelity (a strong bias in movies) and a falling out, which is then reconciled by one of their friends.

The goal of our work is to attempt an understanding of these rich moments of peoples' lives.
Given short clips from a movie, we wish to predict the interactions and relationships, and localize the characters that experience them throughout the movie.
Note that our goals necessitate the combination of visual as well as language cues; some interactions are best expressed visually (\eg~runs with), while others are driven through dialog (\eg~confesses) -- see Fig.~\ref{fig:intro}.
% Fig.~\ref{fig:intro} presents a few examples of interactions and relationships between characters from the movie \emph{Juno}.
As our objectives are quite challenging, we make one simplifying assumption - we use \emph{trimmed} (temporally localized) clips in which the interactions are known to occur.
We are interested in studying two important questions:
(i) can learning to jointly predict relationships and interactions help improve the performance of both? and
(ii) can we use interaction and relationship labels at the clip or movie level and learn to identify/localize the pair of characters involved? We refer to this as \emph{weak track prediction}.
% (ii) can we learn to identify the pair of characters involved along with the prediction of interaction and relationship labels 
% (ii) can we use interaction and relationship labels at the clip or movie level and learn to predict them along with identifying the pair of characters involved? Further, we called this \emph{weak} prediction.
A solution for the first question is attempted using a multi-task formulation operating on several clips spanning the common pair of characters, while the second uses a combination of max-margin losses with multiple instance learning (see Sec.~\ref{sec:model}).

% \todo{need a few more references in the intro, maybe discuss briefly works in relationships / interactions in images? look at the liu2019 paper}

% -- main contributions
\vspace{1mm}
\noindent \textbf{Contributions.}
We conduct our study on 51 movies from the recently released MovieGraphs~\cite{vicol2018moviegraphs} dataset (see Sec.~\ref{subsec:dataset}).
The dataset annotations are based on free-text labels and have long tails for over 300 interaction classes and about 100 relationships.
% many of which have less than 5 or 10 instances.
% We trim these down to 101 interactions and 15 relationships by grouping them based on similarity (see Sec.~\ref{subsec:dataset}).
To the best of our knowledge, ours is the first work that attempts to predict interactions and long-term relationships between characters in movies based on visual and language cues.
We also show that we can learn to localize characters in the video clips while predicting interactions and relationships using weak clip/movie level labels without a significant reduction in performance.
% This will allow us to scale annotations to other movies much faster than the time taken to collect the MovieGraphs dataset.

%% file: relwork.tex
\section{Related Work}
\label{sec:relwork}

We present related work in understanding actions/interactions in videos, studying social relationships, and analyzing movies or TV shows for other related tasks.

\vspace{1mm}
\noindent \textbf{Actions and interactions in videos.}
Understanding actions performed by people can be approached in many different ways.
Among them, action classification involves predicting the dominant activity in a short trimmed video clip~\cite{kuehne2011hmdb51, soomro2012ucf101}, while
action localization involves predicting the activity as well as temporal extent~\cite{girdhar2019video_transf, shou2017cdc, xu2017r_c3d}.
An emerging area involves discovering actions in an unsupervised manner by clustering temporal segments across all videos corresponding to the same action class~\cite{alayrac2016unsupervised,kukleva2019unsupervised,sener2018unsupervised}.

Recently, there has been an interest in creating large-scale datasets (millions of clips, several hundred classes) for learning actions~\cite{abu2016youtube, carreira2017kinetics, diba2019huv, goyal2017something, monfort2019moments} but none of them reflect person-to-person (p2p) multimodal interactions where several complex actions may occur simultaneously.
The AVA challenge and dataset~\cite{gu2018ava} is composed of 15 minute video clips from old movies with \textit{atomic} actions such as pose, person-object interactions, and person-person interactions (\eg~talk to, hand wave).
However, all labels are based on a short (3 second) temporal window, p2p actions are not annotated between multiple people, and relationship labels are not available.
Perhaps closest to our work on studying interactions, Alonso~\etal~\cite{alonso2010high_five} predict interactions between two people using person-centered descriptors with tracks.
However, the TV-Human Interactions dataset~\cite{alonso2010high_five} is limited to 4 visual classes in contrast to 101 multimodal categories in our work.
As we are interested in studying intricate multimodal p2p interactions and long-range relationships, we demonstrate our methods on the MovieGraphs dataset~\cite{vicol2018moviegraphs}.
% An important distinction between action recognition datasets (\eg~Kinetics) and MovieGraphs and AVA is that the latter exhibit a long tail of rare class labels making the tasks extremely challenging, but at the same time revealing the data distribution of classes in the real world.

Recognizing actions in videos requires aggregation of spatio-temporal information.
Early approaches include hand-crafted features such as interest points~\cite{laptev2003stip} and Improved Dense Trajectories~\cite{wang2013idt}.
% As an ImageNet[] counterpart for pretraining models Kinetics dataset~\cite{carreira2017kinetics} is used for videos.
With end-to-end deep learning, spatio-temporal 3D Convolutional Neural Networks (\eg~I3D~\cite{carreira2017kinetics}) are used to learn video representations resulting in state-of-the-art results on video understanding tasks.
% without end-to-end training.
For modeling long-videos, learning aggregation functions~\cite{girdhar17action_vlad, miech17loupe}, subsampling frames~\cite{wang2016tsn}, or accumulating information from a feature bank~\cite{wu2019long} are popular options.
% Recently, \cite{wu2019long} proposed long-term feature banks to accumulate appropriate information from the entire clip. 

\vspace{1mm}
\noindent \textbf{Relationships in still images.}
Most studies on predicting social relationships are based on images~\cite{gallagher2009understanding,goel2019end,li2017dual,socher2013reasoning,sun2017domain}. % li2018visual
For example, the People in Photo Albums (PIPA)~\cite{zhang2015pipa} and the People in Social Context (PISC) datasets~\cite{li2017dual} are popular among social relationship recognition.
The latter contains 5 relationship types (3 personal, 2 professional), and \cite{li2017dual} employs an attention-based model that looks at the entire scene as well as person detections to predict relationships.
Alternately, a domain based approach is presented by Sun~\etal~\cite{sun2017domain} that extends the PIPA dataset and groups 16 social relationships into 5 categories based on Burgental's theory~\cite{burgental2000}.
Semantic attributes are used to build interpretable models for predicting relationships~\cite{sun2017domain}.

We believe that modeling relationships requires looking at long-term temporal interactions between pairs of people, something that still image works do not allow.
Thus, our work is fundamentally different from above literature.

\vspace{1mm}
\noindent \textbf{Social understanding in videos.}
Understanding people in videos goes beyond studying actions.
% action classification~\cite{zhao2019bayesian, choi2012unified, feichtenhofer2019slowfast, feichtenhofer2016convolutional, wang2016tsn},
% joint recognition and temporal localization of events~\cite{shou2017cdc, girdhar2019video_transf, xu2017r_c3d},
% object/person detection and tracking~\cite{feichtenhofer2019slowfast},
Related topics include
clustering faces in videos~\cite{jin2017erdosclustering,tapaswi2019ballclustering},
naming tracks based on multimodal information~\cite{nagrani2017sherlock,tapaswi2012bbt},
studying where people look while interacting~\cite{fan2019understanding,marin2019laeo},
predicting character emotions~\cite{dhall2013emotiw,vicol2018moviegraphs},
modeling spatial relations between objects and characters~\cite{laxton2007leveraging,sun2018actor,zhang2019gnnava},
recognizing actions performed in groups~\cite{bagautdinov2017socialscene,deng2016structure},
predicting effects for characters~\cite{zhou2018movie4d},
producing captions for what people are doing~\cite{rohrbach2017lsmdc,damen2018epickitchens},
answering questions about events, activities, and character motivations~\cite{lei2018tvqa,tapaswi2016movieqa,kim2016pororoqa},
reasoning about social scenes and events~\cite{zadeh2019social_iq,zellers2019vcr},
understanding social relationships~\cite{liu2019social,vicol2018moviegraphs},
and meta-data prediction using multiple modalities~\cite{cascante2019moviescope}.

Perhaps most related to our work on predicting relationships are~\cite{liu2019social,lv2018mmm}.
Lv~\etal~\cite{lv2018mmm} present the first dataset for modeling relationships in video clips, and propose a multi-stream model to classify 16 relationships.
More recently, Liu~\etal~\cite{liu2019social} propose a graph network to capture long-term and short-term temporal cues in the video.
% Pyramid Graph Convolution Network to reason over spatial comprehensive graph in temporal domain to disclose social relationships where both rely on the theory that social relations predictive about the visual attributes and behaviors
Different from above works, we address predicting relationships between pairs of characters in an entire movie.
We propose a joint model for interactions and relationships as they may influence each other, and also localize the characters in the video.

%% file: model.tex
\section{Model}
\label{sec:model}

In this section, we present our approach towards predicting the interactions and relationships between pairs of characters (Sec.~\ref{subsec:intrel}), and localizing characters in the video as tracks (Sec.~\ref{subsec:tracks}).
% Finally, we present the neural architectures used to represent video clips and fusion methods in Sec.~\ref{subsec:architectures}.

\vspace{1mm}
\noindent \textbf{Notation.}
We define $\mcA$ as the set of all interaction labels, both visual and spoken (\eg~runs with, consoles);
and $\mcR$ as the set of all relationship labels between people (\eg~parent-child, friends).
We process complete movies, where each movie $\mcM$ consists of three sets of information:
\vspace{-1mm}
\begin{enumerate}
\itemsep-1mm
\item Characters $\mcC_M = \{c_1, \ldots, c_P\}$, each $c_j$ representing a cluster of all face/person tracks for that character.
\item \emph{Trimmed} video clips annotated with interactions\\ $\mcT_M = \{(v_1, a^*_1, c_{1j}, c_{1k}), \ldots, (v_N, a^*_N, c_{Nj}, c_{Nk})\}$, where
$v_i$ corresponds to a multimodal video clip,
%combining visual scene information, dialogs, and person tracks,
$a^*_i \in \mcA$ is a directed interaction label, and
$c_{ij}$ is used to denote the tracks for character $c_j$ in the clip $v_i$.
\item Directed relationships between all pairs of characters $\mcR_M = \{r^i_{jk} = \mathrm{relationship}(v_i, c_j, c_k)\}$ for all clips $i\in[1,N]$.
% \forall i\in[1,N], j\in[1,P], k\in[1,P], j\neq k \}$.
For simplicity of notation, we assign a relationship label $r^i_{jk}$ to each clip.
However, note that relationships typically span more than a clip, and often the entire movie (\eg~parent-child).
\end{enumerate}
\vspace{-1mm}

For each clip $v_i$, our goal is to predict the primary interaction $a_i$, the characters $c_{ij}$ and $c_{ik}$ that perform this interaction, and their relationship $r^i_{jk}$.
In practice, we process several clips belonging to the same pair of characters as predicting relationships with a single short clip can be quite challenging, and using multiple clips helps improve performance.
% to obtain strong representations for predic
% \note{Actually, we predict relationship on more than one clip.}

We denote the \emph{correct} pair of characters in a tuple $(v_i, a_i^*, c_{ij}, c_{ik})$ from $\mcT$ as $p_i^* = (c_{ij}, c_{ik})$,
and the set of all character pairs as $\mcP_M = \{(c_j, c_k) \forall j,k,j\neq k\}$.

Note that the interaction tuples in $\mcT$ may be temporally overlapping with each other.
For example, Jack may \emph{look at} Jill while she \emph{talks to} him.
We deal with such interaction labels from overlapping clips in our learning approach by masking them out in the loss function.

% For a new movie, our goal of this work is to jointly predict the INTs $\mcA$ and RELs $\mcR$ between characters and .
% For simplicity, we assume that relationships do not change in a movie 
% (\todo{THINK:} how to deal with changes? one option is to call that part a different movie; get some stats to understand this better).
% \noteAnna{we discussed this moment. for simplicity we decided to call the part with unchanged relationships as a movie p.s. will collect statistic}

% The key contribution/novelty is in modeling short-time interactions and long-time relationships jointly, and seeing how they can help in predicting each other.
% We first present our model for understanding interactions in a clip, followed by analyzing the entire movie.

\begin{figure}
\centering
\includegraphics[width=0.9\linewidth]{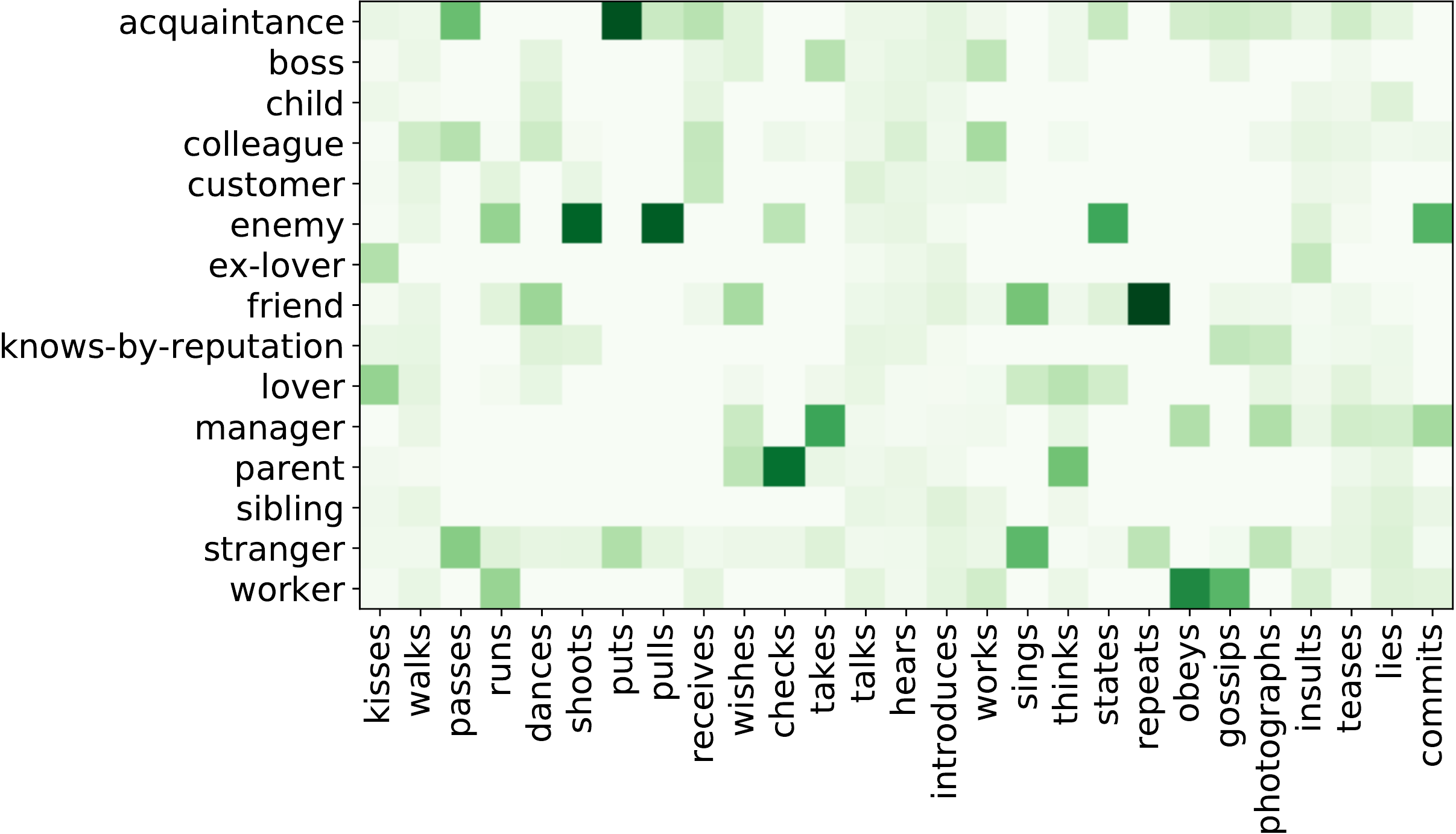}
\vspace{-4mm}
\caption{Normalized correlation map between (selected) interactions and relationships.
Darker regions indicate higher scores.}
\label{fig:heatmap}
\vspace{-4mm}
\end{figure}

\begin{figure*}
\centering
\includegraphics[width=\linewidth]{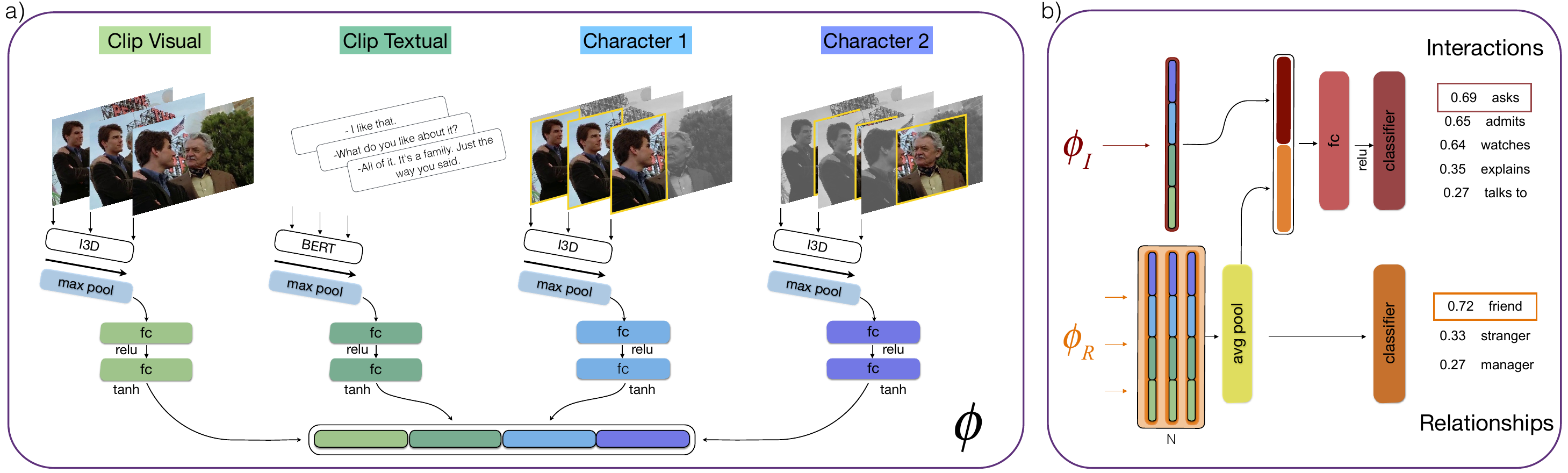}
\vspace{-0.5cm}
\caption{Model architecture.
{\bf Left}: Our input is a trimmed video clip for one interaction, and consists of visual frames and all dialogues within its duration.
Each interaction is associated with two characters, and they are represented visually by extracting features from cropped bounding boxes.
Modalities are processed using fixed pre-trained models (BERT for textual, I3D for visual) to extract clip representations denoted by $\Phi(v)$.
{\bf Right}: In the second panel, we show the architecture of our joint interaction and relationship prediction model.
In particular, multiple clips are used to compute relationships, and we fuse these features while computing interaction labels.}
\label{fig:model}
\vspace{-0.4cm}
\end{figure*}

\subsection{Interactions and Relationships in a Clip}
\label{subsec:intrel}
% \vspace{1mm}
% \noindent \textbf{Motivation.}
Fig.~\ref{fig:heatmap} shows example correlations between a few selected interactions and all 15 relationships in our dataset.
We observe that interactions such as \emph{obeys} go together with \emph{worker-manager} relationships, while an \emph{enemy} may \emph{shoot}, or \emph{pull} (a weapon), or \emph{commit} a crime.
Motivated by these correlations, we wish to learn interactions and relationships jointly.
% We propose models to exploit such correlations in our data.

When the pair of characters that interact is known, we predict their interactions and relationships using a multi-modal clip representation $\Phi(v_i, p_i^*) \in \real^D$.
As depicted in Fig.~\ref{fig:model}, $\Phi$ combines features from multiple sources such as visual and dialog cues from the video, and character representations by modeling their spatio-temporal extents (via tracking).

\vspace{1mm}
\noindent \textbf{Interactions.}
We use a two-layer MLP with a classification layer $\bW^{I2} \in \real^{|\mcA| \times D}, \bb^{I2} \in \real^{|\mcA|}$ to predict interactions between characters.
The score for an interaction $a$ in clip $v$ is computed as
\begin{equation}
\label{eq:s-i}
s_I(v, a) = \sigma_I(\bw^{I2}_a \cdot \mathrm{ReLU}(\bW^{I1} \Phi(v, p^*) + \bb^{I1}) + b^{I2}_a),
\end{equation}
where $\sigma(\cdot)$ represents the sigmoid operator.
We learn the clip representation parameters along with the MLP by minimizing the max-margin loss function for each sample
\begin{equation}
\label{eq:l-i}
L_I(v) = \sum_{\mathclap{\substack{\bar{a} \in \mcA \setminus \mcO_I(v) \\ \bar{a} \neq a^*}}} \;
\left[ m_I - s_I(v, a^*) + s_I(v, \bar{a}) \right]_+ \, ,
\end{equation}
where $[\cdot]_+$ is a ReLU operator, $m_I$ is the margin, and $\mcO_I(v)$ is the set of interaction labels from clips temporally overlapping with $v$.
The loss encourages our model to associate the correct interaction $a^*$ with the clip $v$, while pushing other non-overlapping interaction labels $\bar{a}$ away.
During inference, we predict the interaction for a clip $v$ as $\hat{a} = \argmax_a s_I(v, a)$.

\vspace{1mm}
\noindent \textbf{Relationships.}
While interactions are often short in duration (few seconds to a minute), relationships in a movie may last for several minutes to the entire movie.
To obtain robust predictions for relationships, we train a model that observes several trimmed video clips that portray the same pair of characters.
Let us denote $V_{jk} \subset \{v_1, \ldots, v_N\}$ as one such subset of clips that focus on characters $c_j$ and $c_k$.
In the following, we drop the subscripts $jk$ for brevity.

Similar to predicting interactions, we represent individual clips of $V$ using $\Phi$, apply a pooling function $g(\cdot)$ (\eg~avg, max) to combine the individual clip representations as $\Phi(V, p^*) = g(\Phi(v, p^*))$ and adopt a linear classifier $\bW^R \in \real^{|\mcR| \times D}, \bb^R \in \real^{|\mcR|}$ to predict relationships.
The scoring function
\begin{equation}
\label{eq:s-r}
s_R(V, r) = \sigma_r \left( \bw^R_r \Phi(V, p^*) + b^R_r \right)
\end{equation}
computes the likelihood of character pair $p^*$ from the clips $V$ having relationship $r$.
We train model parameters using a similar max-margin loss function
\begin{equation}
\label{eq:l-r}
L_R(V) = \sum_{\mathclap{\substack{\bar{r} \in \mcR \\ \bar{r} \neq r^*}}} \; 
\left[ m_R - s_R(V, r^*) + s_R(V, \bar{r}) \right]_+ \, ,
\end{equation}
that attempts to score the correct relationship $r^*$ higher than others $\bar{r}$.
Unlike $L_I$, we assume that a single label applies to all clips in $V$.
If a pair of characters change relationships (\eg~from \emph{strangers} to \emph{friends}), we select the set of clips $V$ during which a single relationship is present.
At test time, we predict the relationship as $\hat{r} = \argmax_r s_R(r, V)$.
% \todo{are we doing the evaluation run with sliding window predicting relationship?}

\vspace{1mm}
\noindent \textbf{Joint prediction of interactions and relationships}
is performed using a multi-task formulation.
We consider multiple clips $V$ and train our model to predict the relationship as well as all interactions of the individual clips jointly.
We introduce a dependency between the two tasks by concatenating the clip representations for interactions $\Phi_I(v, p^*)$ and relationships $\Phi_R(V, p^*)$.
Fig.~\ref{fig:model} visualizes the architecture used for this task.
We predict interactions as follows:
\begin{equation}
\label{eq:s-ir}
% \small
s_I(v, V, a) = \sigma(\bw^{I2}_a \cdot \mathrm{ReLU}(\bW^{I1} [\Phi_I(v, p^*); \Phi_R(V, p^*)])).
\end{equation}
Linear layers include biases, but are omitted for brevity.
The loss function $L_I(v)$ now uses $s_I(v, V, a)$, but remains unchanged otherwise.
The combined loss function is
\begin{equation}
\label{eq:l-ir}
L_{IR}(V) = L_R(V) + \frac{\lambda}{|V|} \sum_{v \in V} L_I(v) \, ,
\end{equation}
where $\lambda$ balances the two losses.
% \todo{but it is not true, I do not use all the clips from V.. - implementation detail? :)}

\subsection{Who is interacting?}
\label{subsec:tracks}
Up until now, we assumed that a clip $v$ portrays two \emph{known} characters that performed interaction $a$.
However, movies (and the real world) are often more complex, and we observe that several characters may be interacting simultaneously.
To obtain a better understanding of videos, we present an approach to predict the characters along with the interactions they perform, and their relationship.

While the interaction or relationship may be readily available as a clip-level label, localizing the pair of characters in the video can be a tedious task as it requires annotating tracks in the video.
We present an approach that can work with such \emph{weak} (clip-level) labels, and estimate the pair of characters that may be interacting.

\vspace{1mm}
\noindent \textbf{Predicting interactions and characters.}
As a first step, we look at jointly predicting interactions and the pair of characters.
Recall that $p^*_i$ denotes the \emph{correct} pair of characters in a clip tuple consisting of $v_i$, and $\mcP_M$ is the set of all character pairs in the movie.
% For simplicity, we denote the \emph{correct} pair of characters in a tuple $(v_i, a_i^*, c_{ij}, c_{ik})$ from $\mcT$ as $p_i^* = (c_{ij}, c_{ik})$,
% and the set of all character pairs as $\mcP_M = \{(c_j, c_k) \forall j,k,j\neq k\}$.
We update the scoring function (Eq.~\ref{eq:s-i}) to depend on the chosen pair of characters $p \in \mcP_M$ as
\begin{equation}
\label{eq:s-ic}
% \small
s_{IC}(v, a, p) = \sigma(\bw^{I2}_a \cdot \mathrm{ReLU}(\bW^{I1} \Phi(v, p))) \, ,
\end{equation}
where $\Phi(v, p)$ now encodes the clip representation for any character pair $p$ (we use zeros for unseen characters).
% When the ground-truth pair of characters is known.
We train our model to predict interactions and the character pair by minimizing the following loss
\begin{equation}
\label{eq:l-ic-gt}
\small
L_{IC}(v) = \sum_{\mathclap{\substack{
\bar{a} \in \mcA \setminus \mcO_I(v) \\
\bar{p} \in \mcP_M \\
(\bar{a}, \bar{p}) \neq (a^*\!, p^*)}}} \,
\left[ m_{IC} - s_{IC}(v, a^*\!, p^*) + s_{IC}(v, \bar{a}, \bar{p}) \right]_+ \, .
\end{equation}
If we consider the scoring function $s_{IC}(v, a, p)$ as a matrix of dimensions $|\mcP_M| \times |\mcA|$, the negative samples are taken from everywhere except columns that have an overlapping interaction label $\mcO_I(v)$, and the element where $(\bar{a}=a^*, \bar{p}=p^*)$.
At test time, we compute the character pair prediction accuracy given ground-truth (GT) interaction, interaction accuracy given GT character pair, and joint accuracy by picking the maximum score along both dimensions.

\vspace{1mm}
\noindent \textbf{Training with weak labels.}
When the GT character pair $p^*$ is \emph{not known} during training, we modify the loss from Eq.~\ref{eq:l-ic-gt} by first choosing the pair $\hat{p}^*$ that scores highest for the current parameters and $a^*$, that is known during training.
% {\small
\begin{align}
\label{eq:hatp-ic}
\hat{p}^* &= \arg\max_p s_{IC}(v, a^*\!, p) \, , \\
\label{eq:l-ic-weak}
L^{\mathrm{weak}}_{IC}(v) &= \sum_{\mathclap{\substack{
\bar{a} \in \mcA \setminus \mcO_I(v), \, \bar{a} \neq a^* \\
\bar{p} \in \mcP_M}}}
\left[ m_{IC} - s_{IC}(v, a^*\!, \hat{p}^*) + s_{IC}(v, \bar{a}, \bar{p}) \right]_+ \, .
\end{align}%
% We use a single loss function as the character pair information is incorporated directly into the feature encoding $\Phi$.
In contrast to the case when we know GT $p^*$, we discard negatives from the entire column ($a=a^*$) to prevent minor changes in choosing $\hat{p}^*$ from suppressing other character pairs.
In practice, we treat $s_{IC}(v, a^*\!, p)$ as a multinomial distribution and sample $\hat{p}^*$ from it to prevent the model from getting stuck at only one pair.
Inference is performed in a similar way as above.

\vspace{1mm}
\noindent \textbf{Hard negatives.}
Training a model with max-margin loss can affect performance if the loss is satisfied ($=0$) for most negative samples.
As demonstrated in~\cite{faghri2018vse}, choosing hard negatives by using $\max$ instead of $\sum$ can help improve performance.
We adopt a similar strategy for selecting hard negatives, and compute the loss over all character pairs with their best interaction, \ie~$\sum_{\bar{p} \in \mcP_M} \max_{\bar{a}} (\cdot)$ in Eq.~\ref{eq:l-ic-gt} and~\ref{eq:l-ic-weak}.

\vspace{1mm}
\noindent \textbf{Prediction of interactions, relationships, and characters.}
We present the loss function used to learn a model that jointly estimates which characters are performing what interactions and what are their relationships.
Similar to Eq.~\ref{eq:s-ic}, we first modify the relationship score to depend on $p$:
\begin{equation}
\label{eq:s-rc}
s_{RC}(V, r, p) = \sigma(\bw_r^R g(\Phi(V, p)) + b^R_r) \, .
\end{equation}
This is used in a weak label loss function similar to Eq.~\ref{eq:l-ic-weak}.
{\small
\begin{align}
\label{eq:hatp-rc}
\hat{p}^* &= \arg\,\max_p s_{RC}(V, r^*\!, p) + s_{IC}(v, a^*\!, p) \, , \\
\label{eq:l-rc-weak}
L^{\mathrm{weak}}_{RC}(V) &= \sum_{\mathclap{\substack{
\bar{r} \in \mcR, \, \bar{r} \neq r^* \\
\bar{p} \in \mcP_M}}}
\left[ m_{RC} - s_{RC}(V, r^*\!, \hat{p}^*) + s_{RC}(V, \bar{r}, \bar{p}) \right]_+ \, , \\
L^{\mathrm{weak}}_{IRC}(V) &=  L^{\mathrm{weak}}_{RC}(V) + \frac{\lambda}{|V|} \sum_{v \in V} L^{\mathrm{weak}}_{IC}(v) \, .
\end{align}}

% \lambda^{\mathrm{weak}}_C
During inference, we combine the scoring functions $s_{IC}$ and $s_{RC}$ to produce a 3D tensor in $|\mcP_M| \times |\mcA| \times |\mcR|$.
As before, we compute character pair accuracy given GT $a^*$ and $r^*$,
interaction accuracy given GT $p^*$ and $r^*$, and
relationship accuracy given GT $p^*$ and $a^*$.
We are also able to make joint predictions on all three by picking the element that maximizes the tensor over all three dimensions.

%% file: eval.tex
\section{Experiments}
\label{sec:experiments}

We start by describing implementation details (Sec.~\ref{subsec:data_proc}), followed by a brief analysis of the dataset and the challenging nature of the task (Sec.~\ref{subsec:dataset}).
In Sec.~\ref{subsec:ints_rels} we examine interaction and relationship prediction performance as separate and joint tasks.
Sec.~\ref{subsec:weak_ints_rels} starts with learning interactions and estimating the pair of characters simultaneously.
Finally, we also discuss predicting relationships jointly with interactions and localizing character pairs.
We present both quantitative and qualitative evaluation throughout this section.

\subsection{Implementation Details}

\noindent \textbf{Visual features.}
\label{subsec:data_proc}
We extract visual features for all clips using a ResNeXt-101 model~\cite{hara3dcnns} pre-trained on the Kinetics-400 dataset.
A batch of 16 consecutive frames is encoded, and feature maps are global average-pooled for the clip representation, and average pooled over a region of interest (ROIPool) to represent characters.
Given a trimmed clip $v_i$, we max pool above extracted features over the temporal span of the clip to pick the most important segments.

\vspace{1mm}
\noindent \textbf{Dialog features.}
To obtain a text representation, all dialogues are first parsed into sentences.
A complete sentence may be as short as a single word (\eg~``Hi.'') or consist of several subtitle lines.
Multiple lines are also joined if they end with ``...''.
Then, each sentence is processed using pre-trained BERT-base model with a masked sentence from the next person if it exists.
We supply a mask for every second sentence as done in the sentence pair classification task (for more details, \cf~\cite{devlin2018bert}).
We max pool over all sentences uttered in a trimmed clip to obtain a final representation.

Note that every clip always has a visual representation.
In the absence of dialog or tracks, we set the representations for missing modalities to 0.

\vspace{1mm}
\noindent \textbf{Clip representation} $\Phi$.
We process the feature vector corresponding to each modality obtained after max pooling over the temporal extent with a two-layer MLP.
Dropout (with $p = 0.3$), ReLU and $\tanh(\cdot)$ non-linearities are used in the MLP.
The final clip representation is a concatenation of all modalities (see Fig.~\ref{fig:model} left).

\vspace{1mm}
\noindent \textbf{Multi-label masking.}
As multiple interactions may occur at the same time or have overlapping temporal extents with other clips, we use masking to exclude negative contributions to the loss function by such labels.
$\mcO_I(v)$, the labels corresponding to the set of clips overlapping with $v$, are created by checking for an overlap (IoU) greater than 0.2.

\vspace{1mm}
\noindent \textbf{Learning.}
We train our models with a batch size of 64, and use the Adam optimizer with a learning rate of 3e-5.

\subsection{Dataset}
\label{subsec:dataset}
We evaluate our approach on the MovieGraphs dataset~\cite{vicol2018moviegraphs}.
The dataset provides detailed graph-based annotations of social situations for 7600 scenes in 51 movies.
Two main types of interactions are present---\emph{detailed interactions} (\eg~laughs at) last for a few seconds and are often a part of an overarching \emph{summary interaction} (\eg~entertains) that may span up to a minute.
We ignore this distinction for this work and treat all interactions in a similar manner.
These hierarchical annotations are a common source of multiple labels being associated with the same timespan in the video.

The total number of interactions is different from the number of p2p instances as some interactions involve multiple people.
For example, in an interaction where a couple ($c_j$ and $c_k$) \emph{listens to} their therapist ($c_l$), two p2p instances are created:
$c_j \rightarrow \mathrm{listens\ to} \rightarrow c_l$ and $c_k \rightarrow \mathrm{listens\ to} \rightarrow c_l$.

The dataset is partitioned into train (35 movies), validation (7 movies) and test (9 movies) splits.
The train set consists of 15,516 interactions (and 20,426 p2p instances) and 2,676 pairs of people with annotated relationships.
Validation and test sets have 3,992 and 5,380 p2p instances respectively, and about 600 relationship pairs each.

\vspace{1mm}
\noindent \textbf{Missing labels.}
A relationship label is available for 64\% of the interactions in which at least two people participate.
For a pair of people associated with an interaction, both have track features for 76\% of the dataset.
In other cases one or no characters appear due to failure in tracking or being out of the scene.
For evaluation, we only consider samples that have a relationship, or when a pair of characters appear.

\vspace{1mm}
\noindent \textbf{Merging interaction and relationship labels.}
We reduce the number of interaction labels from 324 to 101,
and relationships from 106 to 15 to mitigate severe problems of long tail with only 1-3 samples per class.
However, the merging does not adversely affect the diversity of classes, \eg~\emph{reassures, wishes, informs, ignores} are different interactions in our label set related to communication.

We adopt a hierarchical approach to merge interactions.
Firstly, all classes are divided into 4 categories:
(i) informative or guiding (\eg~\emph{explains, proposes, assists, guide})
(ii) involving movement (\eg~\emph{hits, plays, embraces, catches});
(iii) neutral valence (\eg~\emph{avoids, pretends, reads, searches}); and
(iv) negative valence (\eg~\emph{scolds, mocks, steals, complains}).
Within each of the subclasses we merge interactions based on how similar their meanings are in common usage -- this process is verified by multiple people.

We also reduce the number of relationships to 15 major classes: stranger, friend, colleague, lover, enemy, acquaintance, ex-lover, boss, worker, manager, customer, knows-by-reputation, parent, child and sibling.

\vspace{1mm}
\noindent \textbf{Directed interactions and relationships}
are used between one person to another.
For example when a \textit{parent} $\rightarrow$ \textit{informs} $\rightarrow$ \textit{child},
the opposite directed interaction from the child to their parent can be \emph{listens to} or \emph{ignores}.
Additionally, interactions and relationships can also be bidirectional, both people act with the same intention~\eg~\textit{lovers} \textit{kiss} each other.

\subsection{Predicting Interactions and Relationships}
\label{subsec:ints_rels}
\input{intr_only_reln_only.tex}

We first present results for predicting interactions and relationships separately, followed by our joint model.

\vspace{1mm}
\noindent \textbf{Interaction classification.}
We analyze the influence of each modality for interaction classification separately in Table~\ref{table:intr_only}.
Dialogs have a stronger impact on model performance as compared to visual features owing to the prevalence of conversation based interactions in movies.
However, both modalities are complementary and when taken together provide a 2.6\% increase in accuracy.
As expected, combining all modalities including tracks for each participating character provide the highest prediction accuracy at 26.1\%.

Apart from accuracy, we report \emph{soft accuracy}, a metric that treats a prediction as correct when it matches any of the interactions overlapping with the clip, \ie~$\hat{a} \in {a^*} \cup \mcO_I(v)$.
When using all modalities, we achieve 32.6\% accuracy.

\begin{figure}
\centering
\includegraphics[width=\linewidth,trim=0mm 2mm 0mm 0mm,clip=true]{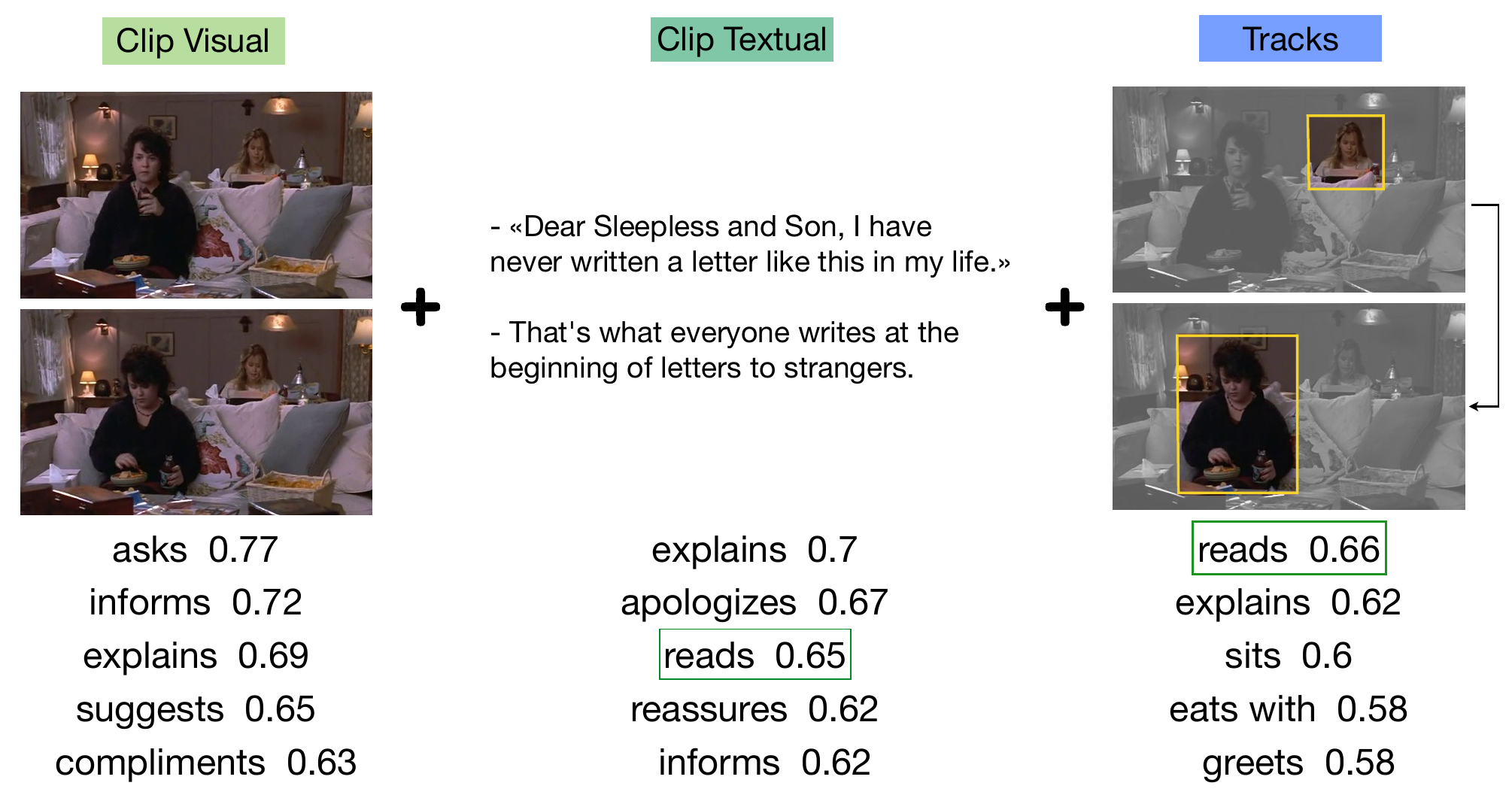} \hfill
% \vspace{mm}

% \par\noindent\rule{\linewidth}{0.4pt}
\includegraphics[width=\linewidth,trim=0mm 2mm 0mm 0mm,clip=true]{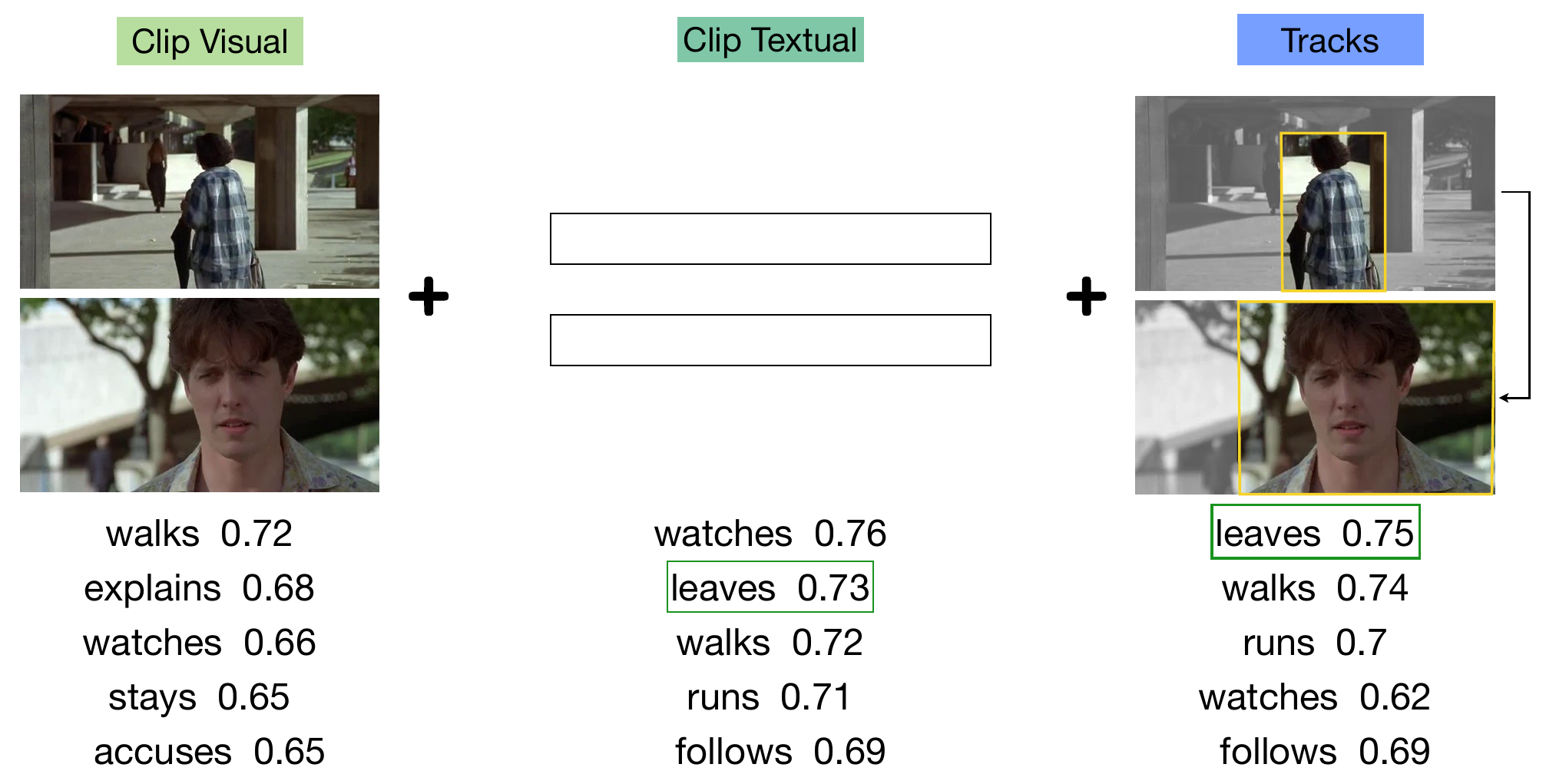}
\vspace{-6mm}
\caption{Influence of different modalities on interaction prediction performance.
In each example, we show the top 5 interactions predicted by the visual cues (left), the visual + dialog cues (center), and visual + dialog + track information (right).
The correct label is marked with a green bounding rectangle.
Discussion in Sec.~\ref{subsec:ints_rels}.}
\label{fig:intr_modality}
\vspace{-4mm}
\end{figure}

In Fig.~\ref{fig:intr_modality} we see two example interactions that are challenging to predict based on visual cues alone.
In the top example, we see that the ground-truth label \emph{reads} is emphasized, possibly due to the dialog mentioning letters, and is chosen with highest score upon examining the visual tracks.
The bottom example is an interesting case where no dialog (all 0 vector) helps predictions.
In this case, our model seems to have learned that \emph{leaving} corresponds to \emph{walking} without any dialog.
Again, by including information about tracks, our model is able to predict the correct label.

We also investigate the influence of different temporal feature aggregation methods in Table~\ref{table:temp_aggr}.
Max-pooling outperforms both average and sum
as it allows to form the clip-level representations including the most influential segments. 

\begin{SCfigure}[1.0][b]
\centering
\includegraphics[width=0.51\linewidth]{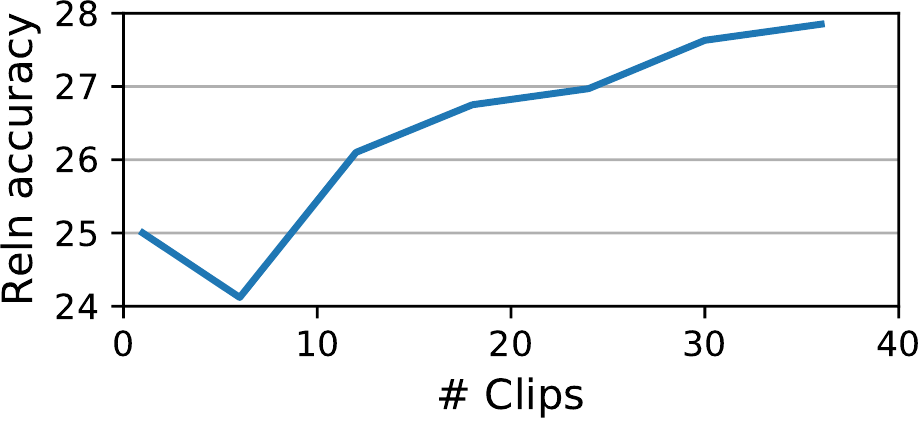}
\caption{Relationship accuracy increases as we analyze more clips. We choose 18 clips as a trade-off between performance and speed.}
\label{fig:n_clips}
\end{SCfigure}

\vspace{1mm}
\noindent \textbf{Relationship classification.}
Relationships are often consistent for long durations in a movie.
For example, \textit{strangers} do not become \textit{friends} in one moment, and \textit{parents} always stay \textit{parents}.
We hypothesize that it is challenging to predict a relationship by watching one interaction, and show the impact of varying the number of clips (size of $V$) in Fig.~\ref{fig:n_clips}.
We see that the relationship accuracy improves steadily as we increase the number of clips.
The drop at 6 clips is within variance.
We choose 18 clips as a trade-off between performance and speed.
During training, we randomly sample up to 18 clips for the same pair of people having the same relationship.
At test time, the clips are fixed and uniformly distributed along all occurrences of pairs of character.

\input{joint_intr_reln_multiclip.tex}
\input{ablations_ints_rels}

\vspace{1mm}
\noindent \textbf{Joint prediction for interactions and relationships.}
% In all our experiments
We set the loss trade-off parameter $\lambda=1.5$ and jointly optimize the network to predict interactions and relationships.
We evaluate different options on how the two tasks are modeled jointly in Table~\ref{table:connections}.
Overall, concatenating relationship feature for predicting interactions performs best (Rel. $\rightarrow$ Int.).
% Recall that the relationships are used to inform the scores for interactions (Eq.~\ref{eq:s-ir}).
Table~\ref{table:joint_intr_reln_multiclip} shows that the relationship accuracy improves by 1.3\%, while interactions gain a meagre 0.2\%.

On further study, we observe that some interactions achieve large improvements, while others see a drop in performance.
For example, interactions such as \emph{hugs} (+17\%), \emph{introduces} (+14\%), and \emph{runs} (+12\%), 
are associated with specific relationships:
\emph{siblings, child, lover} with \emph{hugs}; \emph{enemy, lover} with \emph{runs}.
On the other hand, a few other interactions such as \emph{talks to, accuses, greets, informs, yells} see a drop in performance from 1-8\%, perhaps as they have the same top-3 relationships: \emph{friend, colleague, stranger}.

Relationships show a similar trend.
\emph{Sibling, acquaintance, lover} correspond to specific interactions such as \emph{hugs, greets, kisses} and improve by 11\%, 8\%, and 7\% respectively.
While \emph{boss} and \emph{manager} have rather generic interactions \emph{asks, orders, explains} and reduce by 5-7\%.

We observe that joint learning does helps improve performance.
However, interactions performed by people with common relationships, or relationships that exhibit common interactions are harder for our joint model to identify leading to small overall improvement.
We believe this is made harder due to the long tail classes.

\subsection{Localizing Characters}
\label{subsec:weak_ints_rels}
\input{intr_tracks.tex}

We present an evaluation of character localization and interaction prediction in Table~\ref{table:intr_tracks}.
We report \textbf{interaction} accuracy given the correct character pair;
\textbf{character} pair prediction accuracy given the correct interaction; and
the overall accuracy as \textbf{joint}.

\vspace{1mm}
\noindent \textbf{Training with full supervision.}
In the case when the pair of characters are known (ground-truth pair $p^*$ is given), we achieve 25.5\% accuracy for interactions.
This is comparable to the setting where we only predict interactions (at 26.1\%).
We believe that the difference is due to our goal to maximize the score for the correct interaction and character pair over the entire matrix $|\mcP_M| \times |\mcA|$.
The joint accuracy is 14.2\%, significantly higher than random at 0.15\%.

\vspace{1mm}
\noindent \textbf{Training with weak supervision.}
% In this setup, we consider a weak label scenario.
Here, interaction labels are applicable at the clip-level, and we are unaware of which characters participate in the interaction even during training.
Table~\ref{table:intr_tracks} shows that sampling a character pair is better than $\argmax$ in Eq.~\ref{eq:hatp-ic} (4.6\% vs. 7.8\% joint accuracy) as it prevents the model from getting stuck at a particular selection.
Furthermore, switching training from sum over all negatives to hard negatives (sum-max) after a burn-in period of 20 epochs further improves accuracy to 8.2\%.

\begin{figure}
\centering
\includegraphics[width=\linewidth]{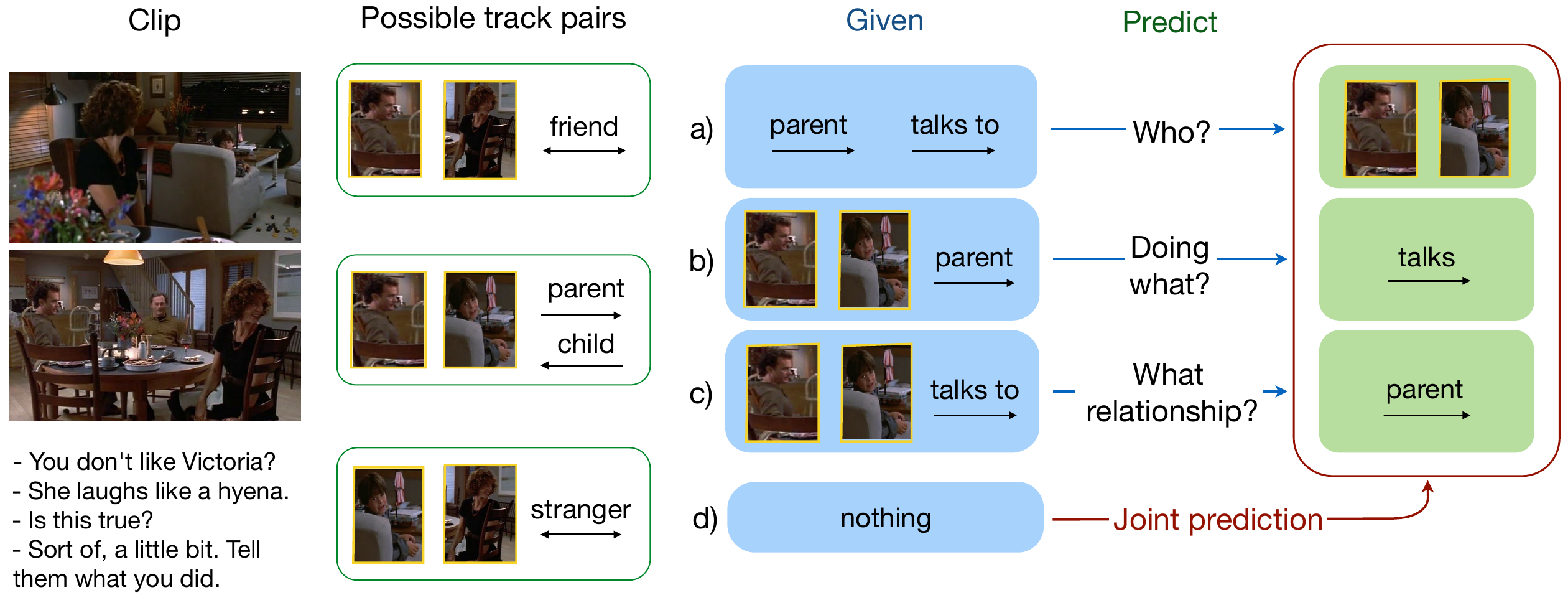}
\vspace{-6mm}
\caption{Example for joint prediction of interaction (Int), relationship (Rel), and character pair (Char) by our model.
The visual clip, dialog, and possible track pairs are presented on the left.
Given 2 pieces of information, we are able to answer the third:
\emph{Who?} Int + Rel $\rightarrow$ Char;
\emph{Doing what?} Char + Rel $\rightarrow$ Int; and
\emph{What relationship?} Char + Int $\rightarrow$ Rel.
We can also jointly predict all three components by maximizing scores along all dimensions of the 3D tensor.
Best seen on screen with zoom.}
\label{fig:weak_tracks}
\vspace{-3mm}
\end{figure}

\input{intr_tracks_reln.tex}

\vspace{1mm}
\noindent \textbf{Joint character localization, interaction and relationship prediction}
is presented in Table~\ref{table:intr_tracks_reln}.
In the case of learning with GT character pairs (fully supervised), including learning of relationships boosts accuracy for predicting character pairs to an impressive 88.3\%.
The interaction accuracy also increases to 25.8\% as compared against 25.5\% when training without relationships (Table~\ref{table:intr_tracks}).

When learning with weak labels, we see similar trends as before.
Both multinomial sampling and switching from all (sum) to hard (sum-max) negatives improves the joint accuracy to a respectable 2.14\% as compared to 2.71\% with full supervision.
Fig.~\ref{fig:weak_tracks} shows an example prediction from our dataset.
We present joint prediction when no information is provided in part d in contrast to parts a, b, c where two of three pieces of information are given.

\vspace{1mm}
\noindent \textbf{Test set.}
\input{test_set.tex}
Table~\ref{table:test_set} compiles results of all our models on the test set.
We see similar trends, apart from a drop in relationship accuracy due to different val and test distributions.

Overall, we observe that learning interactions and relationships jointly helps improve performance, especially for classes that have unique correspondences, but needs further work on other categories.
Additionally, character localization is achievable and we can train models with weak labels without significant drop in performance.

%% file: intr_only_reln_only.tex
%!TEX root = ../main.tex
\begin{table}[!t]
\centering
\small
\tabcolsep=0.15cm
\vspace{-1mm}
\begin{tabular}{ccc ccc}
\toprule
\multicolumn{3}{c}{Modalities} & \multicolumn{3}{c}{Interaction Accuracy} \\
Visual & Dialog & Tracks & Top-1 & Top-1 Soft & Top-5 \\
\midrule
\cmark &   -    &   -    & 18.7 & 24.6 & 45.8 \\
  -    & \cmark &   -    & 22.4 & 30.1 & 50.6 \\
\cmark & \cmark &   -    & 25.0 & 31.9 & 54.8 \\
\cmark & \cmark & \cmark & \textbf{26.1} & \textbf{32.6}  & \textbf{57.3} \\
\bottomrule
\end{tabular}
\vspace{-2mm}
\caption{Interaction prediction accuracy for different modalities.}
\vspace{-5mm}
\label{table:intr_only}
\end{table}

% \begin{table}[!t]
% \centering
% \begin{tabular}{c c cc}
% \toprule
% \multirow{2}{*}{\# Clips} & \multirow{2}{*}{Pooling} & \multicolumn{2}{c}{Reln Accuracy} \\
% & & Top-1 & Top-3 \\
% \midrule
%   1    &   -    & 22.1 & 52.6 \\
%   6    &  avg   & 24.5 & 57.0 \\
%   6    &  max   & 25.4 & 51.9 \\
% \bottomrule
% \end{tabular}
% % \vspace{1mm}
% \caption{all modalities: text, visual, tracks | mid fusion model | relationships only}
% \vspace{-2mm}
% \label{table:reln_only}
% \end{table}

%% file: joint_intr_reln_multiclip.tex
%!TEX root = ../main.tex
\begin{table}[!t]
\centering
\small
\begin{tabular}{c c cc c}
\toprule
Task  & Random & Int. only & Rel. only & Joint \\
\midrule
Interaction  & 0.99 & 26.1 &  -   & \textbf{26.3} \\
Relationship & 6.67 &  -   & 26.8 & \textbf{28.1} \\
\bottomrule
\end{tabular}
% \begin{tabular}{c c cc}
% \toprule
% Method  & Fusion & Int. & Rel. \\
% \midrule
% Random    & -  &  0.99   & 6.67 \\
% \midrule
% Int. only   &   -     & 26.1 &  - \\
% Rel. only   &   -    &  -   & 26.8 \\
% \midrule
% % Joint 1  & \cmark & 27.0 & 22.8 \\
% Joint  & \cmark & \textbf{26.3} & \textbf{28.1} \\
% % Joint       &  6  & best & \cmark & xx.x & xx.x \\
% \bottomrule
% \end{tabular}
\vspace{-2mm}
\caption{Top-1 accuracy for the joint prediction of Int. and Rel.}
\vspace{-3mm}
\label{table:joint_intr_reln_multiclip}
\end{table}

%% file: ablations_ints_rels.tex
%!TEX root = ../main.tex
\begin{table}[!t]
\centering
\small
\begin{minipage}{.48\linewidth}
\centering
\begin{tabular}{c cc}
\toprule
Method  & Int. & Rel. \\
\midrule
Rel. $\leftrightarrow$ Int.   & 25.3 & 26.8 \\
Rel. $\leftarrow$ Int.   & \textbf{26.3} & 25.9 \\
Rel. $\rightarrow$ Int.   & \textbf{26.3} & \textbf{28.1} \\
\bottomrule
\end{tabular}
\vspace{-2mm}
\caption{Different architectures for joint modeling of interactions and relationships.}
% Options for the connections of latent representations of interactions and relationships for the joint prediction of both. Rel. $\rightarrow$ Int. corresponds to the right part of the Fig.~\ref{fig:model}.}
\vspace{-2mm}
\label{table:connections}
\end{minipage}\hfill
\begin{minipage}{.42\linewidth}
\centering
\begin{tabular}{c c}
\toprule
Method  & Int. \\
\midrule
avg   & 24.2 \\
sum   & 25.4 \\
max   & \textbf{26.1} \\
\bottomrule
\end{tabular}
\vspace{-2mm}
\caption{Impact of temporal aggregation methods on interaction accuracy.}
\vspace{-2mm}
\label{table:temp_aggr}
\end{minipage} 
\end{table}

%% file: intr_tracks.tex
%!TEX root = ../main.tex
\begin{table}[!t]
\centering
\small
\tabcolsep=1.5mm
\begin{tabular}{c cc ccc}
\toprule
\multirow{2}{*}{Supervision} & \multirow{2}{*}{Negatives} & Multinom. & \multicolumn{3}{c}{Accuracy} \\
 & & Sampling & Int. & Character & Joint \\
\midrule
Random    & - & - & 0.99  &  15.42   & 0.15 \\
\midrule
Full      &   sum   &   -    & 25.5 & 55.0 & 14.2 \\
% GT      &   max   &   -    & 21.6 & 63.3 & 13.9 \\
% GT      & sum-max &   -    & 22.1 & 63.2 & 14.0 \\
\midrule
Weak    &   sum   &   -    & 18.9 & 20.0 &  4.6 \\
% Weak    & sum-max &   -    & 17.1 & 35.2 &  8.7 \\
% Weak    &   max   &   -    & 20.1 & 41.1 &  9.4 \\
Weak    &  sum   & \cmark & 25.1 & 25.0 &  7.8 \\
Weak    & sum-max & \cmark & 23.0 & 32.3 & 8.2 \\
\bottomrule
\end{tabular}
\vspace{-2mm}
\caption{Joint prediction of \textbf{interactions} and \textbf{character pairs} for fully and weakly supervised settings.
See Sec.~\ref{subsec:weak_ints_rels} for a discussion.}
\vspace{-4mm}
\label{table:intr_tracks}
\end{table}

%% file: intr_tracks_reln.tex
%!TEX root = ../main.tex
\begin{table}[!t]
\centering
\small
\tabcolsep=1.2mm
\begin{tabular}{c cc cccc}
\toprule
\multirow{2}{*}{Supervision} & \multirow{2}{*}{Negatives} & Multinom. & \multicolumn{4}{c}{Accuracy} \\
 && Sampling & Int. & Rel. & Char. & Joint \\
\midrule
Random   & - & - & 0.99 & 6.67 &  15.42   & 0.01 \\
\midrule
Full      &   sum      &   -      & 25.8  & 16.6 & 88.3 & 2.71 \\
\midrule
Weak    &   sum      &  \cmark  & 25.8 & 12.0 & 42.0 & 0.86 \\
Weak    &   sum-max  &  \cmark  & 20.8 & 21.8 & 33.9 & 2.14 \\
% Weak    &   sum-max  &  \cmark  & 23.4 & 19.7 & 23.9 & 1.28 \\
\bottomrule
\end{tabular}
\vspace{-2mm}
\caption{Joint \textbf{interaction}, \textbf{relationship}, and \textbf{character pair} prediction accuracy.
Other labels are provided when predicting columns: Int., Rel., and Char.
See Sec.~\ref{subsec:weak_ints_rels} for a discussion.}
\vspace{-2mm}
\label{table:intr_tracks_reln}
\end{table}

%% file: test_set.tex
%!TEX root = ../main.tex
\begin{table}[!t]
\centering
\small
\tabcolsep=1.4mm
\begin{tabular}{l c cccc}
\toprule
\multirow{2}{*}{All methods} & \multirow{2}{*}{Supervision} & \multicolumn{4}{c}{Accuracy} \\
        &              & Int.    & Rel. & Char. & Joint \\
\midrule
Int only & -                & 20.7 &  -   &  -   &  -   \\
Rel only & -                &  -   & 22.4 &  -   &  -   \\
Int + Rel  & -              & 20.7 & 20.5 &  -   &  -   \\
\midrule
Int + Char & Full           & 19.7 &  -   & 52.8 & 11.1 \\
Int + Char & Weak           & 17.9 &  -   & 29.7 & 6.34 \\
\midrule
Int + Rel + Char & Full     & 20.0 & 18.6 & 88.8 & 2.29 \\
Int + Rel + Char & Weak     & 15.6 & 29.6 & 21.6 & 1.50 \\
\bottomrule
\end{tabular}
\vspace{-2mm}
\caption{Evaluation on the test set.
The columns Int., Rel, and Char refer to interaction, relationship, and character pair prediction accuracy.
During joint learning with full/weak supervision, individual accuracies are reported when other labels are given.}
\vspace{-4mm}
\label{table:test_set}
\end{table}

%% file: conc.tex
\section{Conclusion}
\label{sec:conc}

We presented new tasks and models to study the interplay of interactions and relationships between pairs of characters in movies.
Our neural architecture efficiently encoded multimodal information in the form of visual clips, dialog, and character pairs that were demonstrated to be complementary for predicting interactions.
Joint prediction of interactions and relationships was found to be particularly beneficial for some classes.
We also presented an approach to localize character pairs given their interaction/relationship labels at a clip-level, \ie~without character-level supervision during training.
We will share modifications made to the MovieGraphs dataset to promote future work in this exciting area of improving understanding of human social situations.

\vspace{1mm}
\footnotesize{
\noindent \textbf{Acknowledgements.}
This project was partially supported by the Louis Vuitton - ENS Chair on Artificial Intelligence and the French government under Agence Nationale de la Recherche as part of ``Investissements d'avenir'' program, reference ANR-19-P3IA-0001 (PRAIRIE 3IA Institute).
}

%% file: appendix.tex
% \onecolumn

We provide additional analysis of our task and models including confusion matrices, prediction examples for all our models, skewed distribution of number of samples for our classes, and diagrams depicting how we grouped the interaction and relationship classes.

%%%%%%%%%%%%%%%%%%%%%%%%%%%%%%%%%%%%%%%%%%%%%%%%%%%%%%%%%%%%%%%%%%
% INTERACTIONS IMPROVE WITH MODALITIES - EXAMPLES
%%%%%%%%%%%%%%%%%%%%%%%%%%%%%%%%%%%%%%%%%%%%%%%%%%%%%%%%%%%%%%%%%%
\begin{figure*}
\centering
\includegraphics[width=0.8\linewidth]{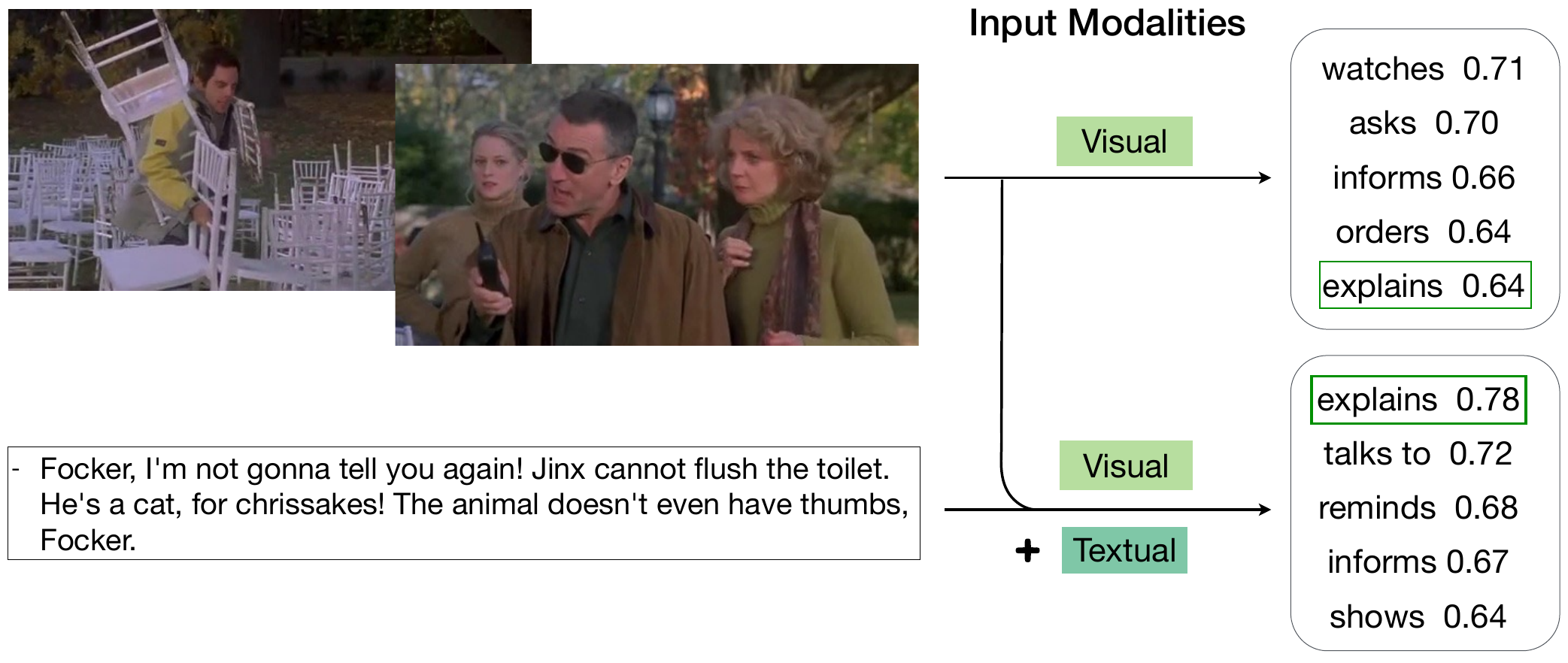}
\vspace{-3mm}
\caption{Improvement in prediction of \textbf{interactions} by including textual modality in addition to visual.
The model learns to recognize subtle differences between interactions based on dialog.
The example is from \textit{Meet the Parents (2000)}.}
\label{fig:v_m}
\vspace{-3mm}
\end{figure*}

\begin{figure*}
\centering
\includegraphics[width=0.8\linewidth]{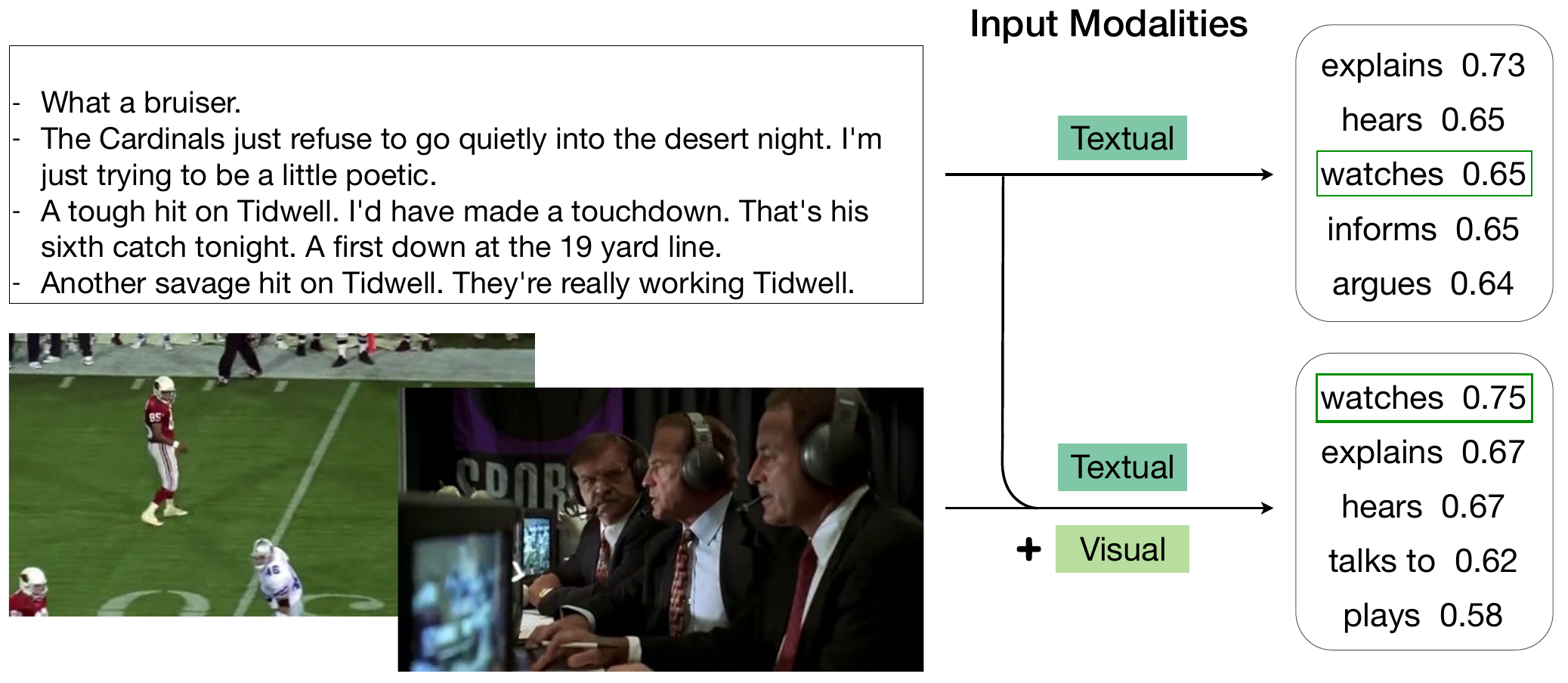}
% \vspace{-6mm}
\caption{Improvement in prediction of \textbf{interactions} by including visual modality in addition to textual.
The top-5 predicted interactions reflect the impact of visual input rather than relying only on the dialog.
The example is from \textit{Jerry Maguire (1996)}.}
\label{fig:t_m}
\vspace{-3mm}
\end{figure*}

\begin{figure*}
\centering
\includegraphics[width=0.8\linewidth]{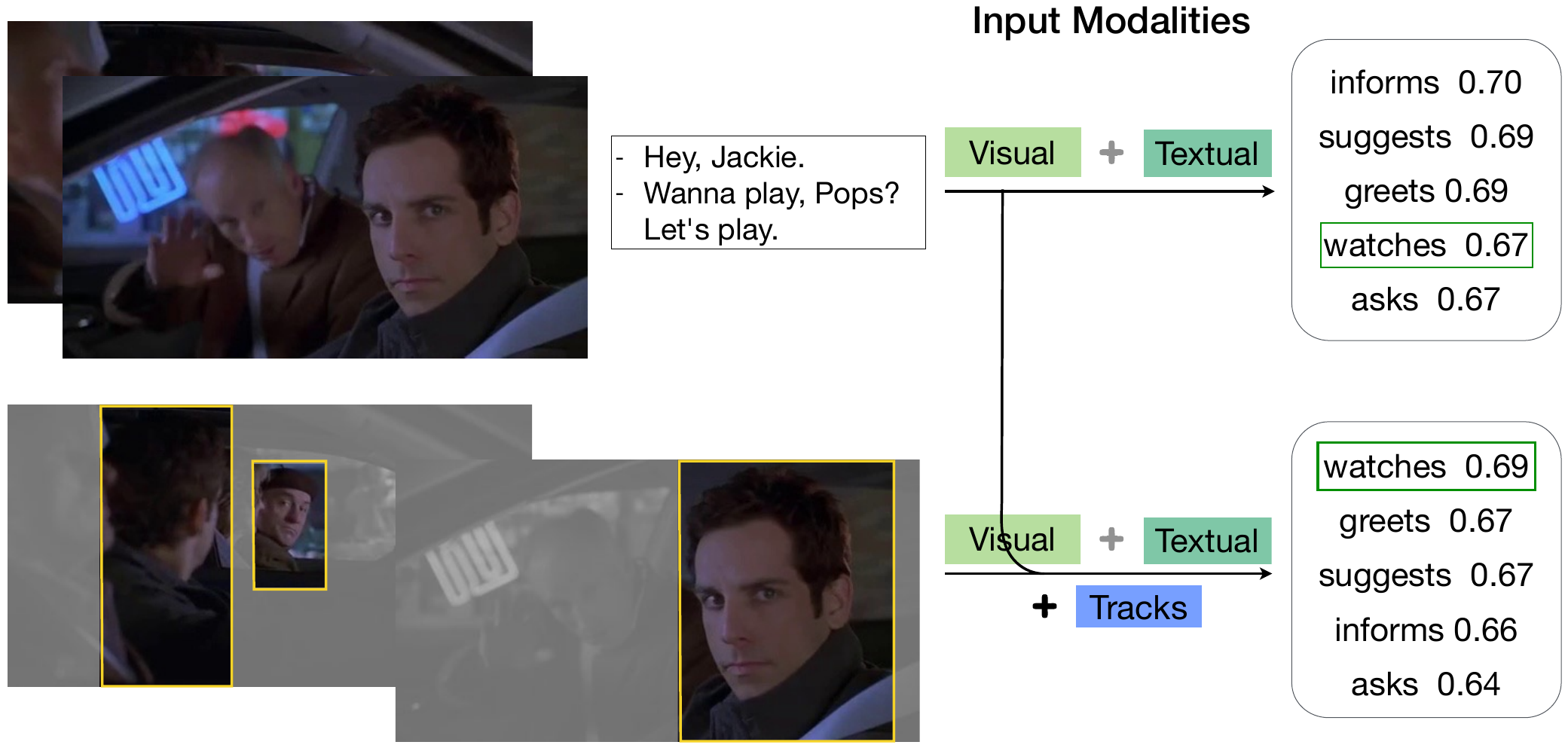}
% \vspace{-6mm}
\caption{Improvement in prediction of \textbf{interactions} by including the pair of tracks modality in addition to visual and textual cues.
The model can concentrate its attention on visual cues for the two people of interest instead of looking only at the clip level.
The example is from \textit{Meet the Parents (2000)}.}
\label{fig:m_tr}
\vspace{-3mm}
\end{figure*}

%%%%%%%%%%%%%%%%%%%%%%%%%%%%%%%%%%%%%%%%%%%%%%%%%%%%%%%%%%%%%%%%%%
% TOP-5 INTERACTIONS WITH IMPROVEMENT DUE TO MODALITIES
%%%%%%%%%%%%%%%%%%%%%%%%%%%%%%%%%%%%%%%%%%%%%%%%%%%%%%%%%%%%%%%%%%
\begin{figure*}
\centering
\includegraphics[width=0.3\linewidth]{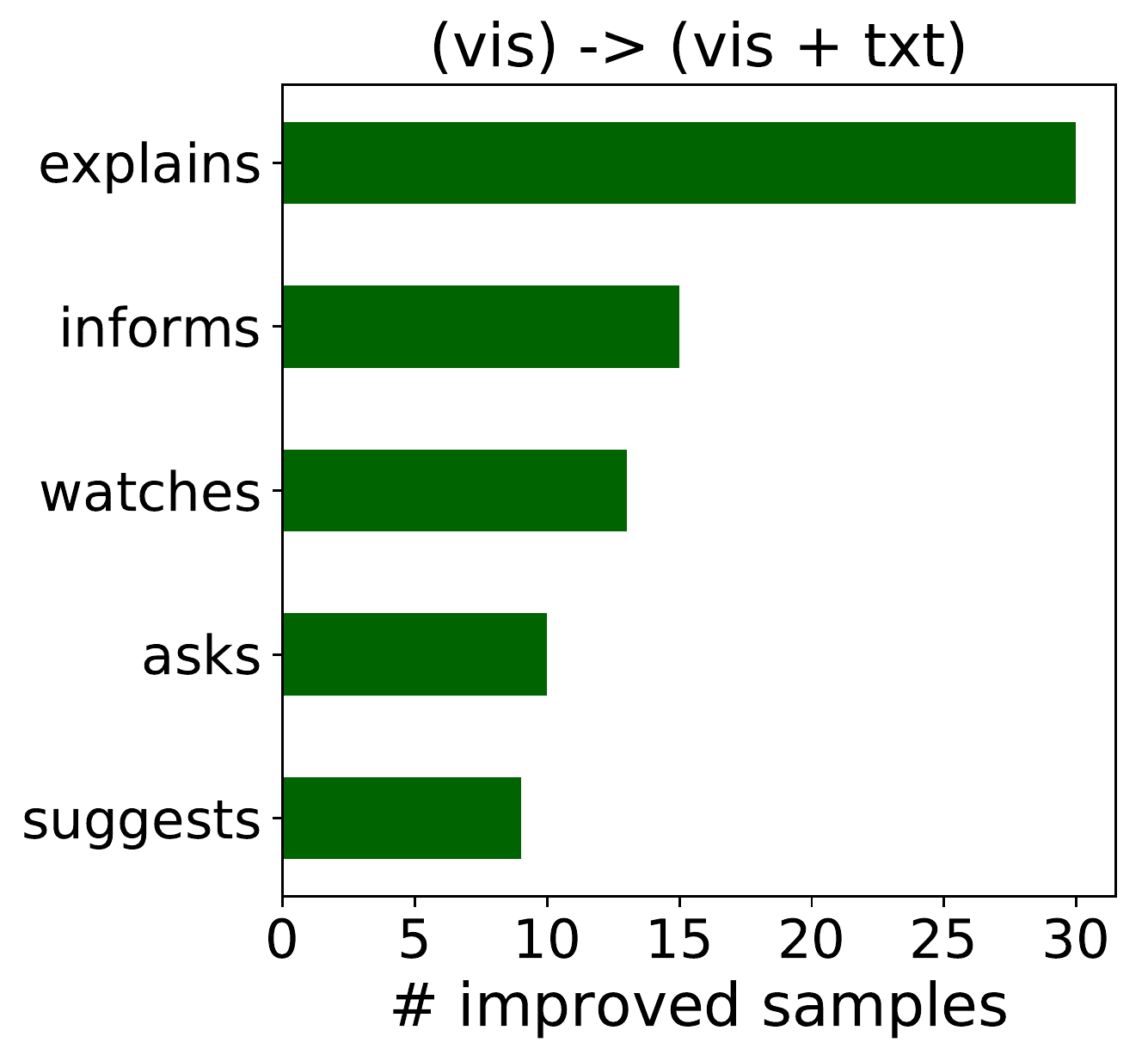} \hfill
\includegraphics[width=0.3\linewidth]{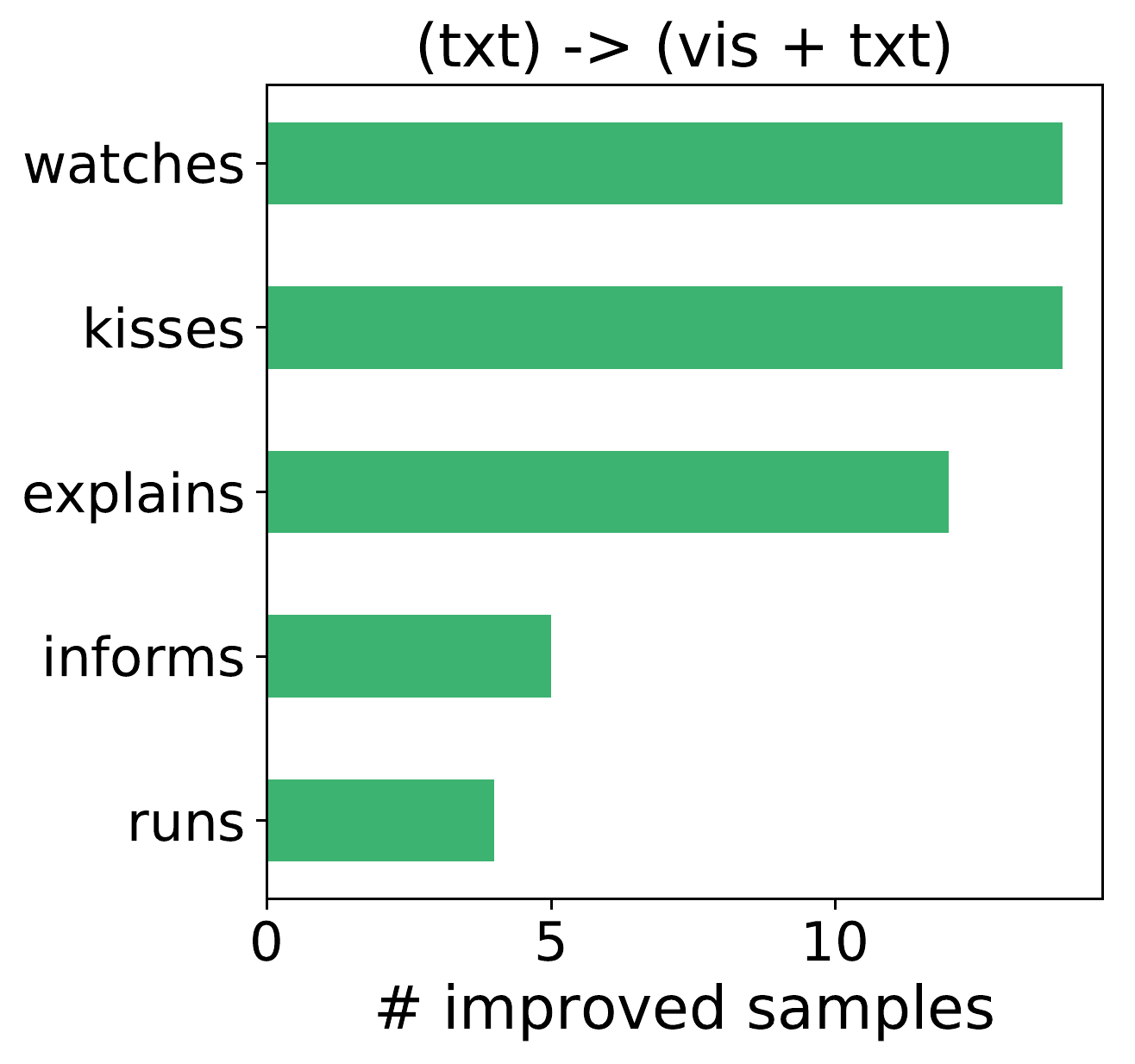} \hfill
\includegraphics[width=0.3\linewidth]{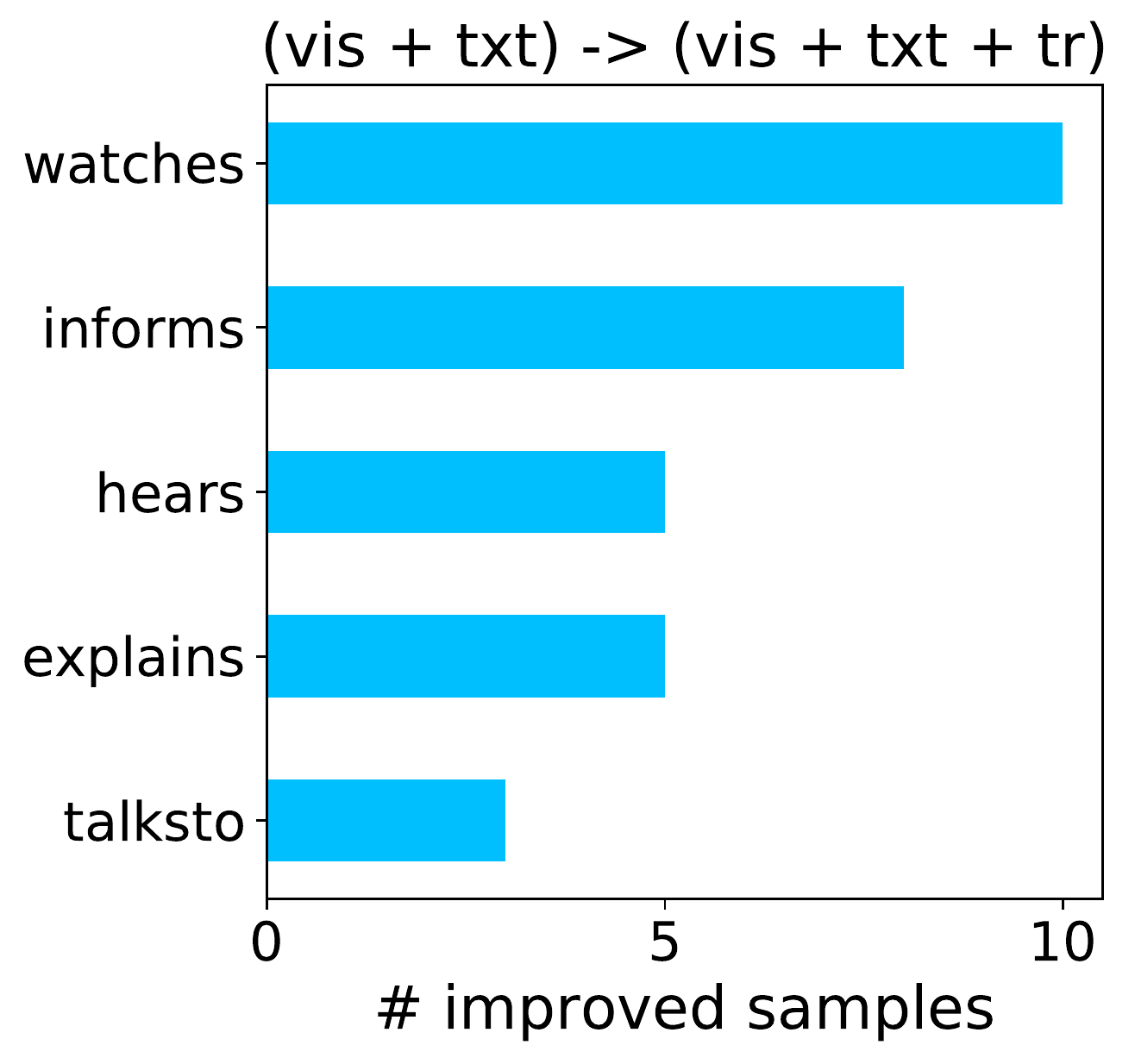}
\caption{Each plot shows 5 \textbf{interaction} classes that have the most number of improved instances by including an additional modality.
Specifically, the $x$-axis denotes the number of samples in which interaction prediction performance improves.
\textbf{Left}: From only \textit{visual} clip representation to \textit{visual and textual}.
As expected, using dialogues in addition to video frames boosts performance for classes that rely on dialog~\eg~\textit{explains, informs}.
\textbf{Middle}: From only \textit{textual} clip representation to \textit{visual and textual}. Visual clip representations influence classes as \textit{kisses, runs} during which people usually do not talk (dialog modality filled with zeros).
\textbf{Right}: Finally, including all three modalities \textit{visual, textual, tracks} improves performance over using \textit{visual and textual}.
Track pair localization improves recognition of classes typically used in group activities.}
\label{fig:top5_modals}
\end{figure*}

%%%%%%%%%%%%%%%%%%%%%%%%%%%%%%%%%%%%%%%%%%%%%%%%%%%%%%%%%%%%%%%%%%
% CONFUSION MATRICES (INTERACTIONS, RELATIONSHIPS)
%%%%%%%%%%%%%%%%%%%%%%%%%%%%%%%%%%%%%%%%%%%%%%%%%%%%%%%%%%%%%%%%%%
\begin{figure*}
\centering
\includegraphics[width=0.37\linewidth]{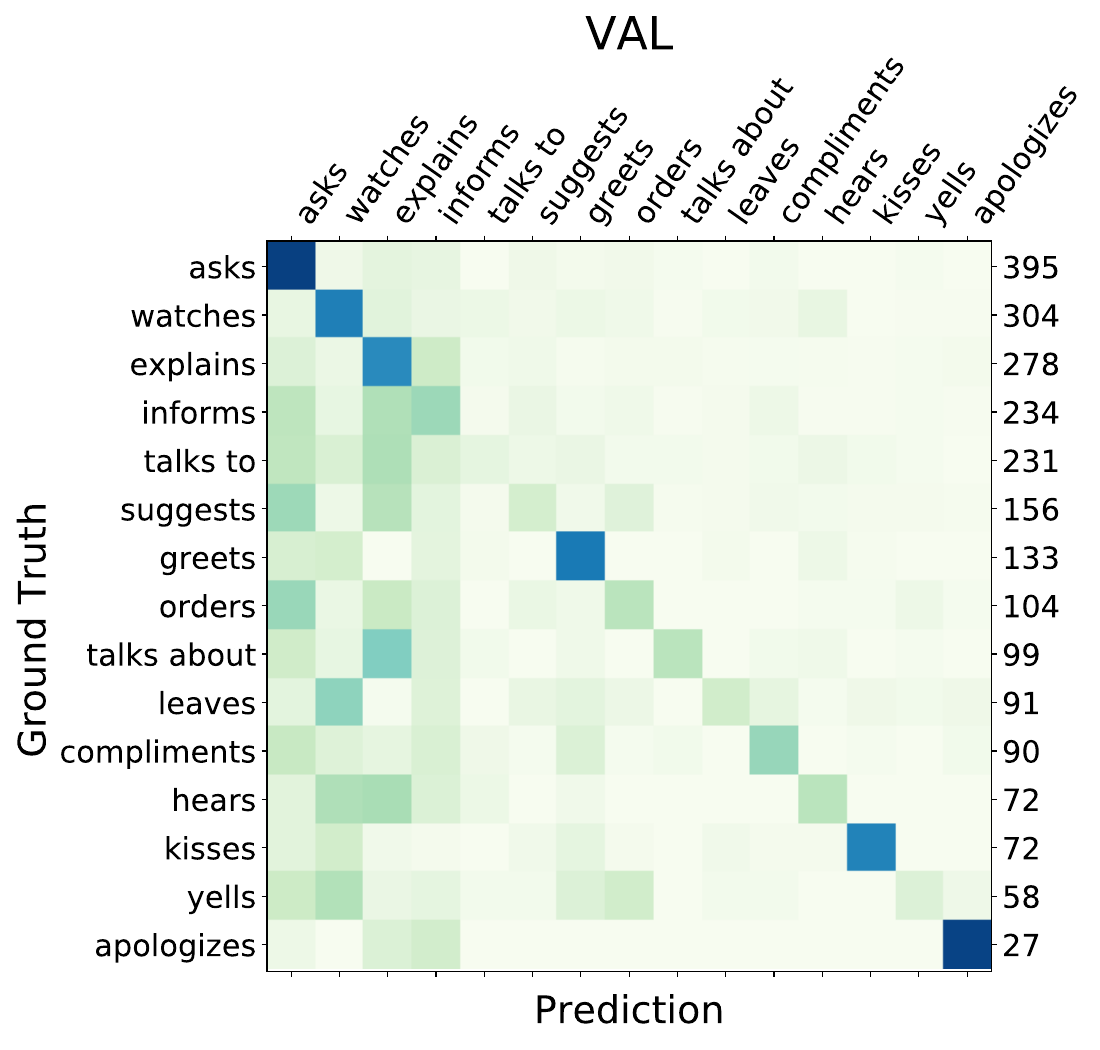} \quad \quad
\includegraphics[width=0.37\linewidth]{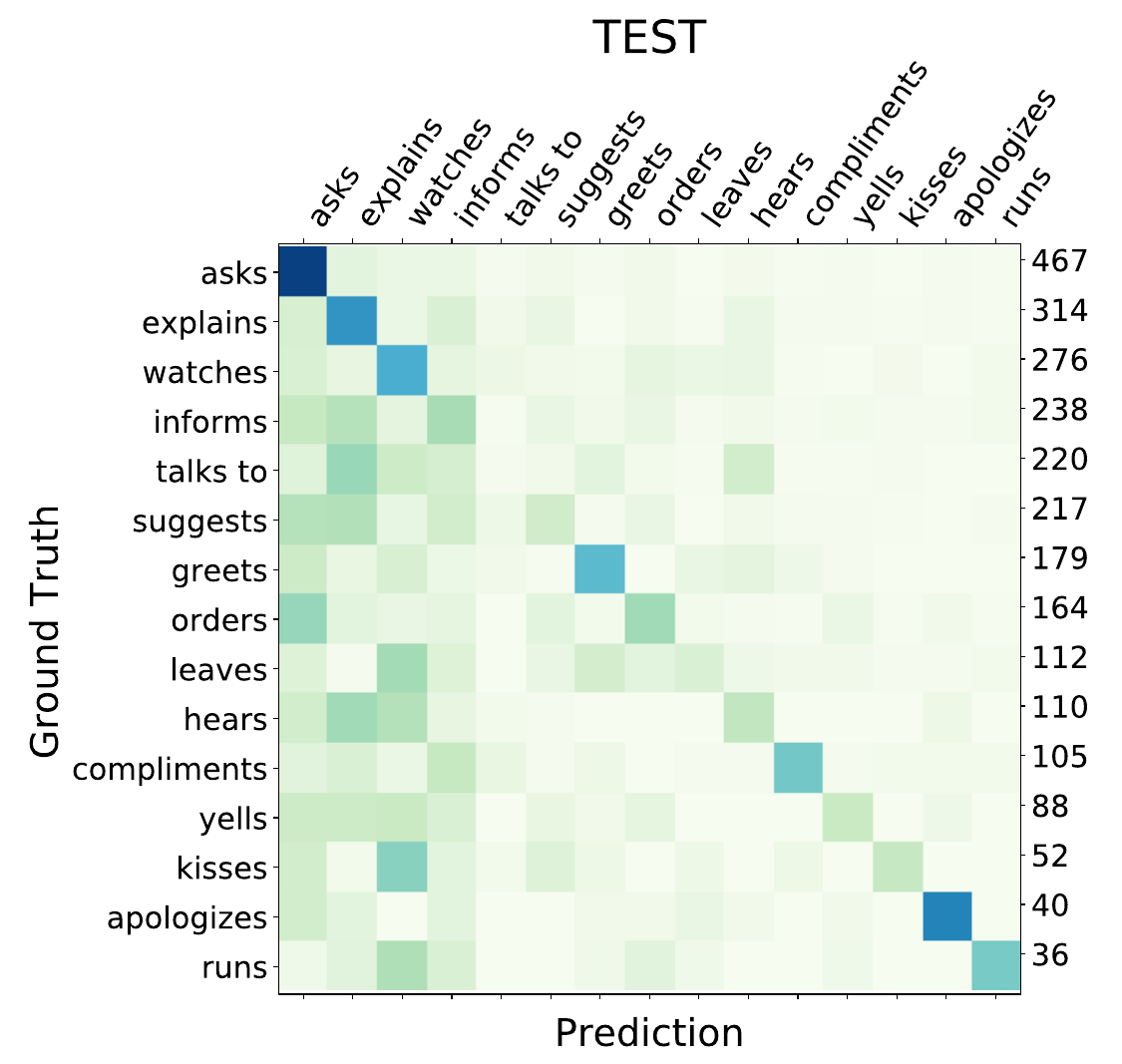}
\vspace{-0.2cm}
\caption{Confusion matrices for top-15 most common interactions for validation set (left) and test set (right).
Model corresponds to the ``Int. only'' performance of 26.1\% shown in Table~\ref{table:joint_intr_reln_multiclip}.
Numbers on the right axis indicate number of samples for each class.}
\label{fig:ints_conf_mat}
% \vspace{-0.2cm}
\end{figure*}

\begin{figure*}
\centering
\vspace{-0.5cm}
\includegraphics[width=0.37\linewidth]{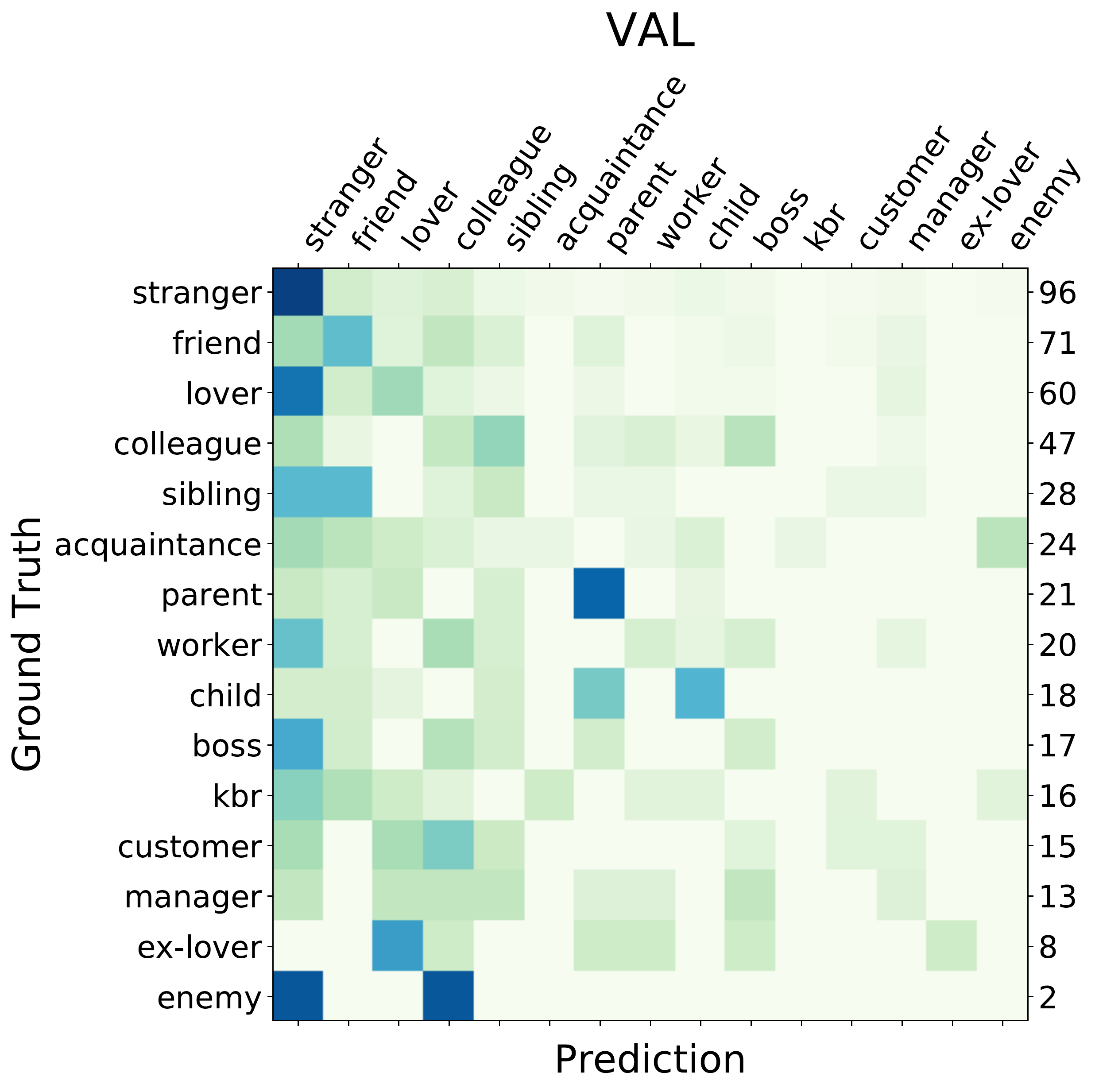} \quad \quad
\includegraphics[width=0.37\linewidth]{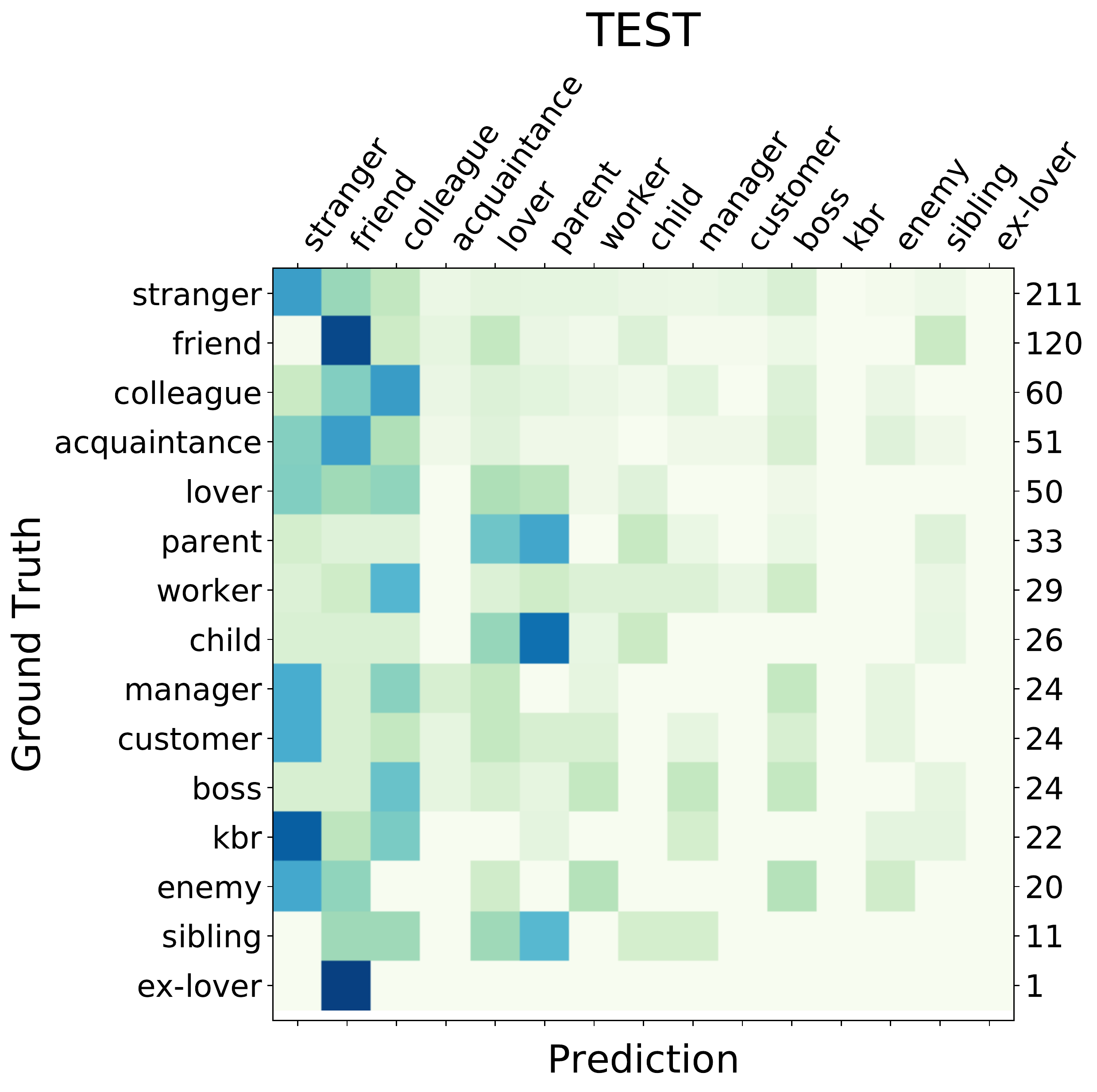}
\vspace{-0.2cm}
\caption{Confusion matrices for all relationships for validation set (left) and test set (right).
Model corresponds to the ``Rel. only'' performance of 26.8\% shown in Table~\ref{table:joint_intr_reln_multiclip}.
Numbers on the right axis indicate number of samples for each class.}
\label{fig:rels_conf_mat}
\vspace{-0.2cm}
\end{figure*}

%%%%%%%%%%%%%%%%%%%%%%%%%%%%%%%%%%%%%%%%%%%%%%%%%%%%%%%%%%%%%%%%%%
% JOINT MODELING IMPROVES INTERACTIONS, RELATIONSHIPS
%%%%%%%%%%%%%%%%%%%%%%%%%%%%%%%%%%%%%%%%%%%%%%%%%%%%%%%%%%%%%%%%%%
\begin{figure*}
\centering
\includegraphics[width=0.8\linewidth,trim=0mm 2mm 0mm 0mm,clip=true]{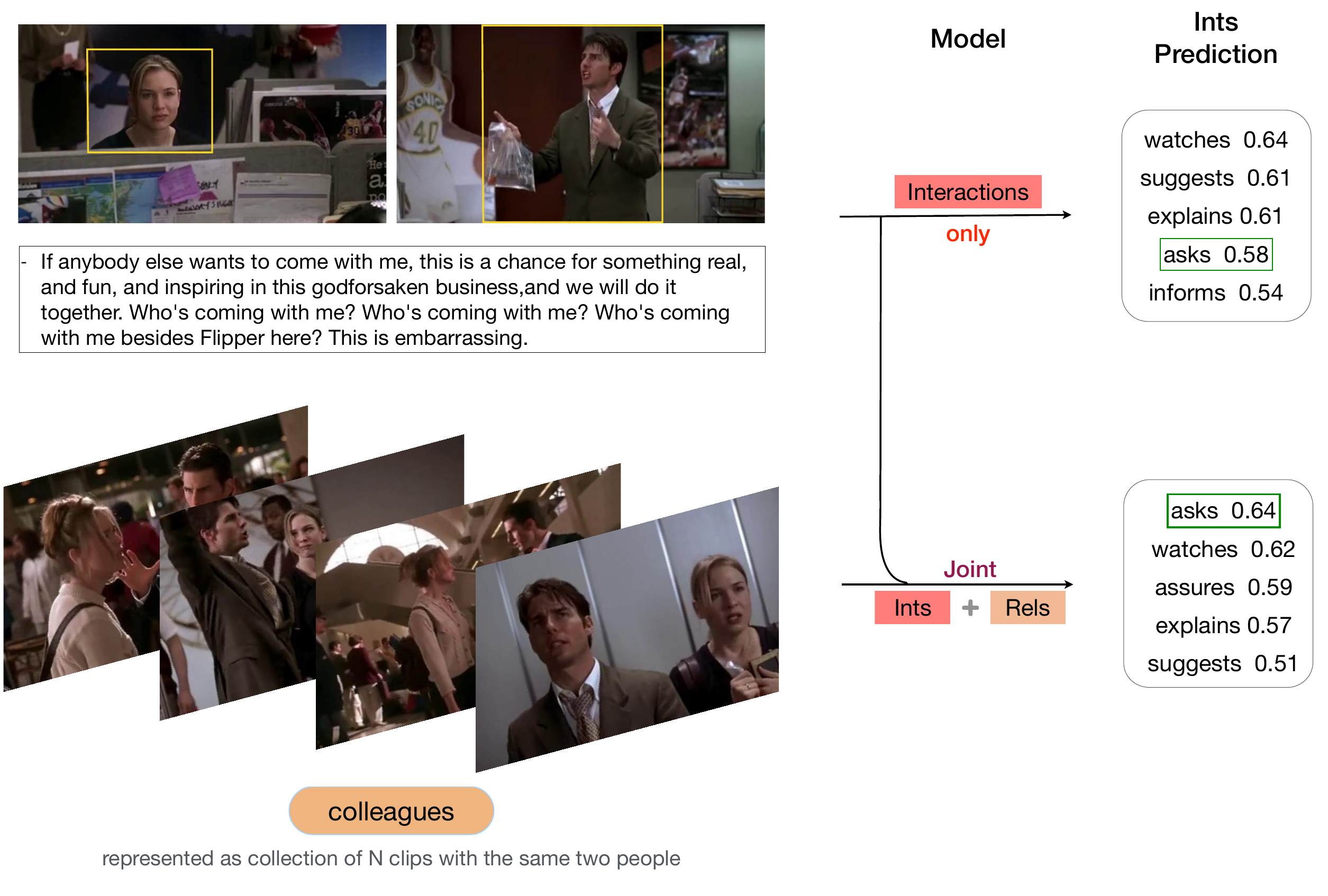} 
\vspace{1cm}
\includegraphics[width=0.8\linewidth,trim=0mm 2mm 0mm 0mm,clip=true]{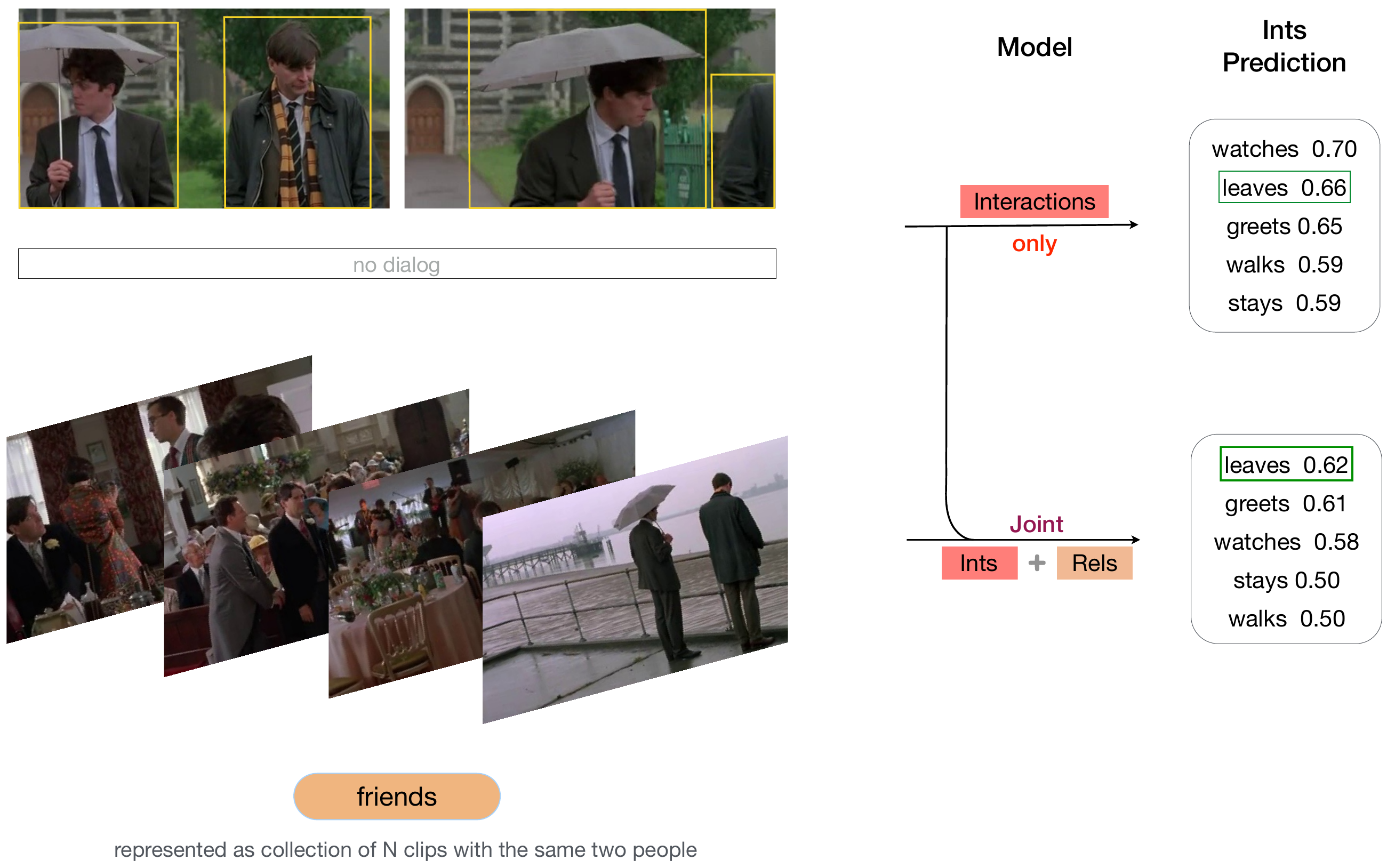}
% \vspace{-6mm}
\caption{We show examples where training to predict interactions and relationships jointly helps improve the performance of \textbf{interactions}.
\textbf{Top}: In the example from \textit{Jerry Maguire (1996)}, the joint model looks at several clips between Dorothy and Jerry and is able to reason about them being \textit{colleagues}.
This in turn helps refine the interaction prediction to \textit{asks}.
\textbf{Bottom:} In the example from \textit{Four Weddings and Funeral (1994)}, the model observes several clips from the entire movie where Charles and Tom are friends, and reasons that the interaction should be \emph{leave} (which contains the \textit{leave together} class).
Note that there is no dialog for this clip.}
\label{fig:intr_mods}
\vspace{-2mm}
\end{figure*}

\begin{figure*}
\centering
\includegraphics[width=\linewidth,trim=0mm 1mm 0mm 0mm,clip=true]{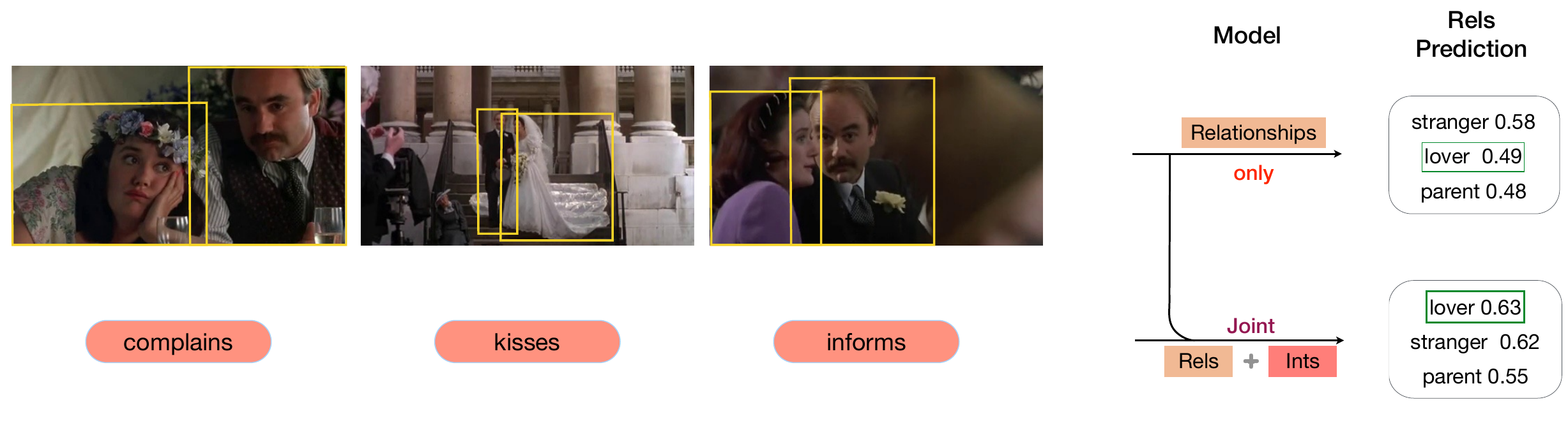} 
\vspace{1cm}
\includegraphics[width=\linewidth,trim=0mm 1mm 0mm 0mm,clip=true]{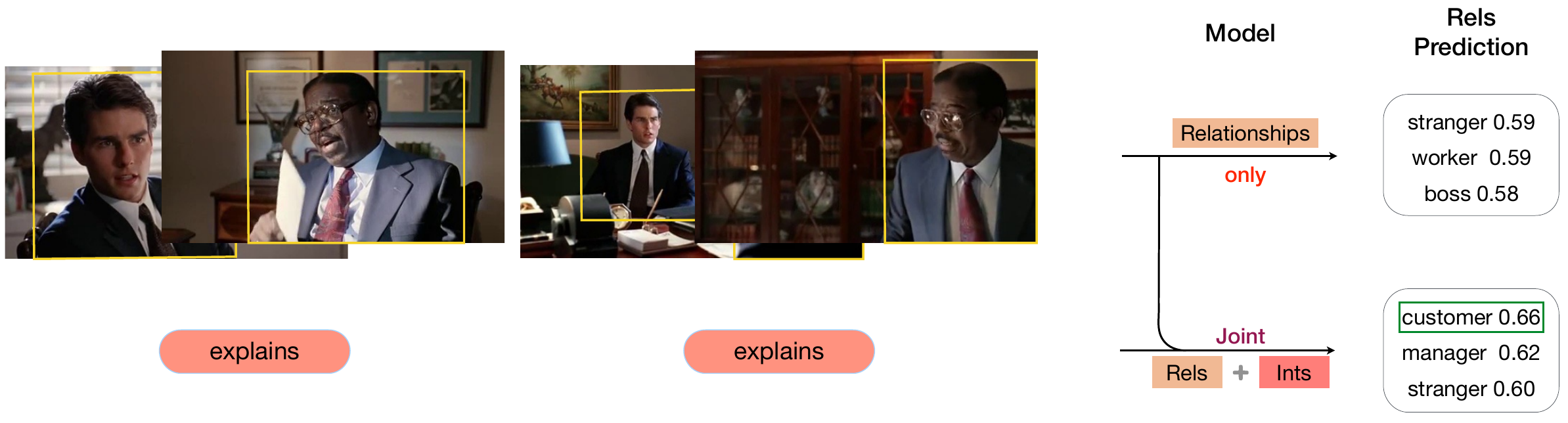}
\vspace{-1cm}
\caption{We show examples where training to predict interactions and relationships jointly helps improve the performance of \textbf{relationships}.
\textbf{Top}: In the movie \textit{Four Weddings and Funeral (1994)}, clips between Bernard and Lydia exhibit a variety of interactions (\eg~kisses) that are more typical between \textit{lovers} than \textit{strangers}.
\textbf{Bottom}: In the movie \textit{The Firm (1993)}, Frank and Mitch meet only once for a consultation, and are involved in two clips with the same interaction label \textit{explains}. 
Our model is able to reason about this interaction, and it encourages the relationship to be \textit{customer} and \textit{manager}, instead of \emph{stranger}.}
\label{fig:rels_mods}
% \vspace{-2mm}
\end{figure*}

%%%%%%%%%%%%%%%%%%%%%%%%%%%%%%%%%%%%%%%%%%%%%%%%%%%%%%%%%%%%%%%%%%
% PREDICTING TRACKS, HELPED BY RELATIONSHIPS
%%%%%%%%%%%%%%%%%%%%%%%%%%%%%%%%%%%%%%%%%%%%%%%%%%%%%%%%%%%%%%%%%%
\begin{figure*}
\centering
\includegraphics[width=\linewidth]{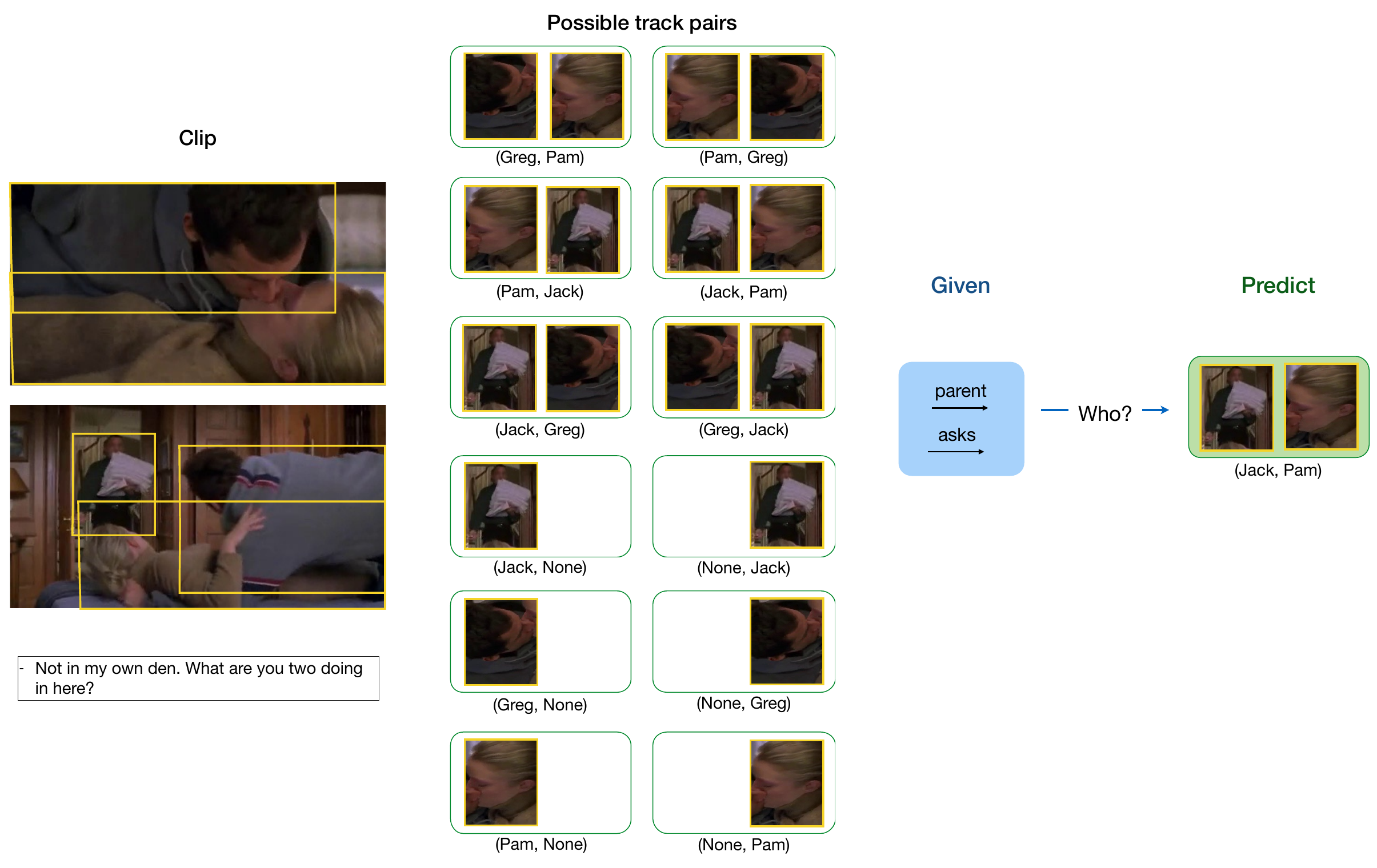}
\vspace{-0.5cm}
\caption{We illustrate an example from the movie \emph{Meet the Parents (2000)} where a father (Jack) walks into a room while his daughter (Pam) and the guy (Greg) are kissing.
Our goal is to predict the two characters when the interaction and relationship labels are provided.
In this particular example, we see that Dad asks Pam a question (What are you two doing in here?). Note that their relationship is encoded as (Pam $\rightarrow$ child $\rightarrow$ Jack), or equivalently, (Jack $\rightarrow$ parent $\rightarrow$ Pam).
% In this particular example, we see that Pam first asks her Dad a question (Who enters without knocking?). Note that their relationship is encoded as (Pam $\rightarrow$ child $\rightarrow$ Jack), or equivalently, (Jack $\rightarrow$ parent $\rightarrow$ Pam).
When searching for the pair of characters with a given interaction \textbf{asks} and relationship as \textbf{parent}, our model is able to focus on the question at the clip level as it is asked by Jack in the interaction, and correctly predict \textit{(Jack, Pam)} as the ordered character pair.
Note that our model not only considers all possible directed track pairs (\eg~(Greg, Pam) and (Pam, Greg)) between characters, but also singleton tracks (\eg~(Jack, None)) to deal with situations when a person is absent due to failure in tracking or does not appear in the scene.}
\label{fig:weak_tracks}
\vspace{-0.4cm}
\end{figure*}

%%%%%%%%%%%%%%%%%%%%%%%%%%%%%%%%%%%%%%%%%%%%%%%%%%%%%%%%%%%%%%%%%%
% JOINT PREDICTION OF EVERYTHING
%%%%%%%%%%%%%%%%%%%%%%%%%%%%%%%%%%%%%%%%%%%%%%%%%%%%%%%%%%%%%%%%%%
\begin{figure*}
\centering
\includegraphics[width=\linewidth]{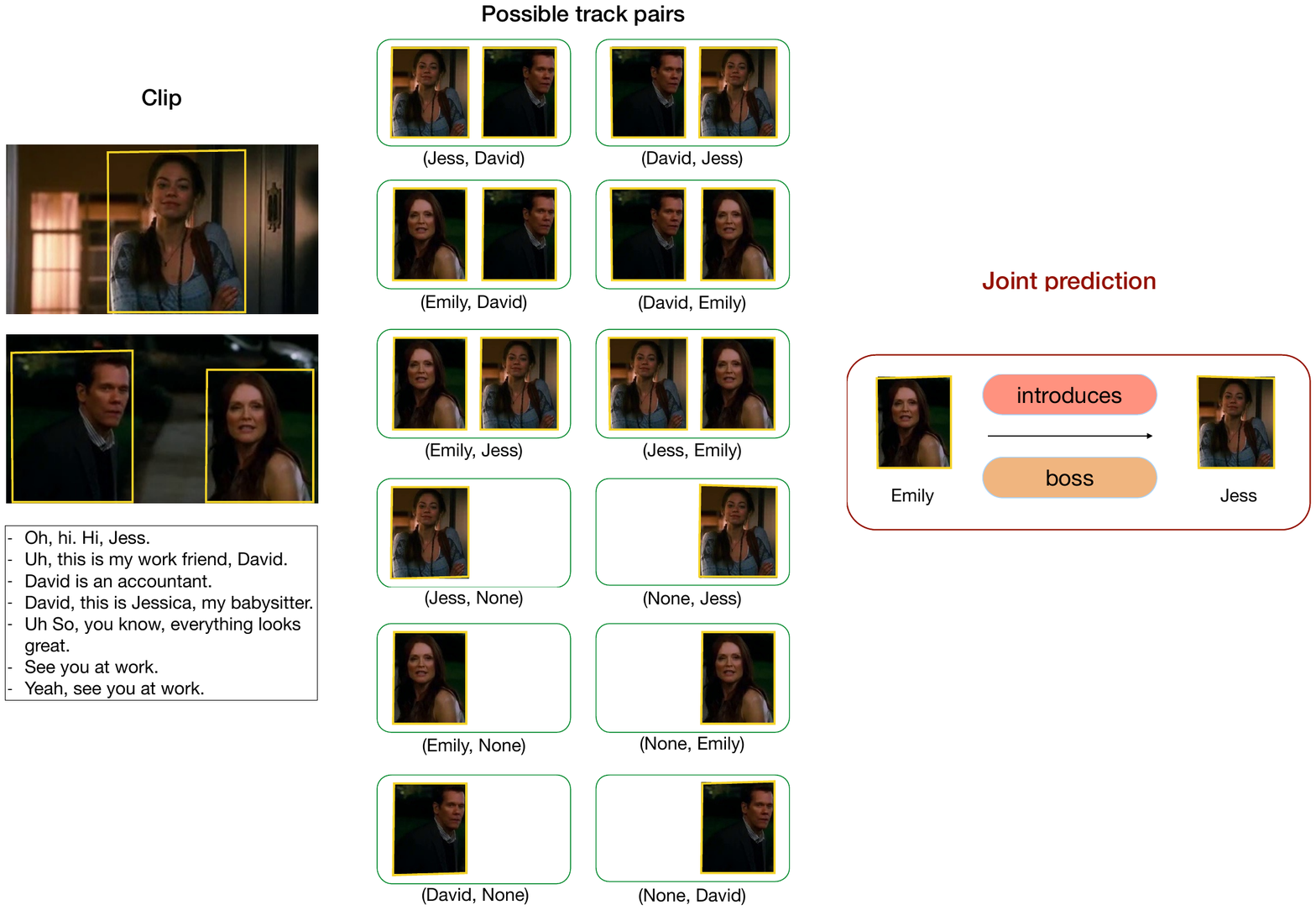}
\vspace{-0.5cm}
\caption{We present an example where our model is able to correctly and jointly predict all three components: track pair, interaction class and relationship type for the clip obtained from the movie \textit{Crazy, Stupid, Love (2011)}. This clip contains three characters which leads to 12 possible track pairs (including singletons to deal with situations when a person is absent due to failure in tracking or does not appear in the scene).
% as constant quantities there are 101 interaction and 15 relationship classes.
The model is able to correctly predict the two characters, their order, interaction and relationship.
In this case, \textit{Emily} \textbf{introduces} David \textbf{to} \textit{Jess}.
Jess is also her hired babysitter, and thus their relationship is -- \emph{Emily} is \textbf{boss} of \emph{Jess}.}
\label{fig:weak_tracks_joint}
\vspace{-0.4cm}
\end{figure*}

%%%%%%%%%%%%%%%%%%%%%%%%%%%%%%%%%%%%%%%%%%%%%%%%%%%%%%%%%%%%%%%%%%
% DISTRIBUTIONS (INTERACTIONS, RELATIONSHIPS)
%%%%%%%%%%%%%%%%%%%%%%%%%%%%%%%%%%%%%%%%%%%%%%%%%%%%%%%%%%%%%%%%%%
\begin{figure*}
\centering
\vspace{2cm}
\includegraphics[width=\linewidth]{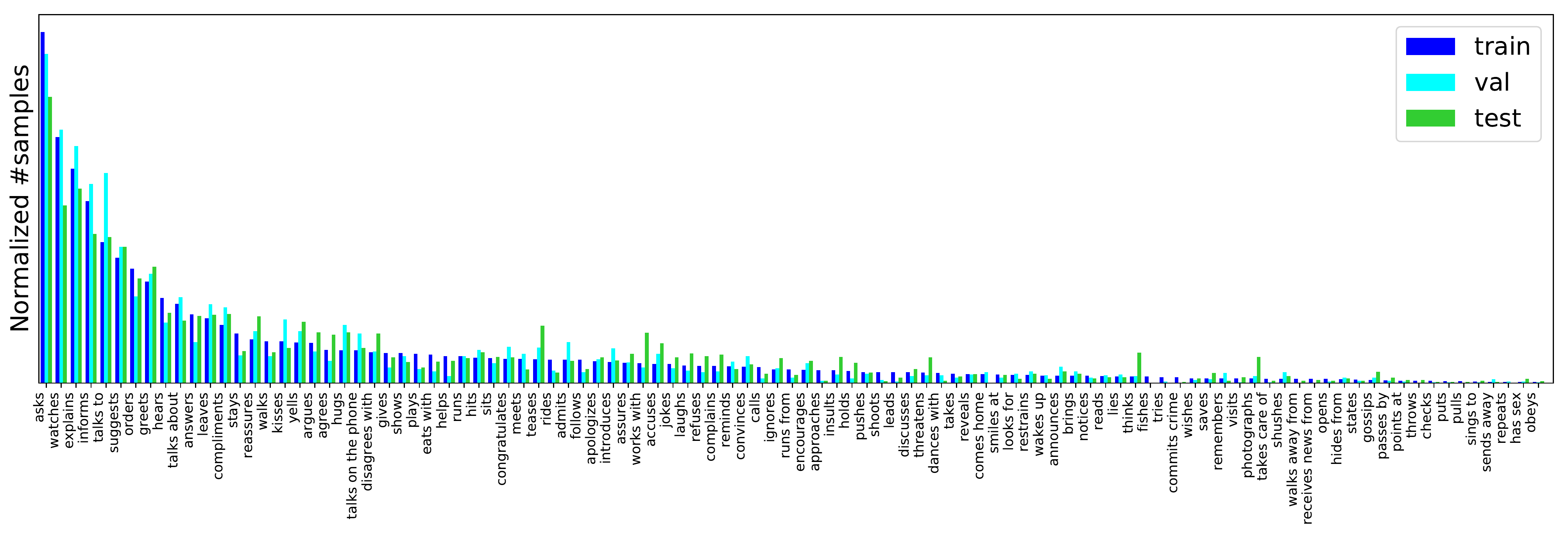}
\vspace{-0.5cm}
\caption{Distribution of interaction labels in train/val/test sets.
Sorted by descending order based on train set.}
\label{fig:ints_distr}
\vspace{1cm}
\end{figure*}

\begin{figure*}
\centering
\includegraphics[width=\linewidth]{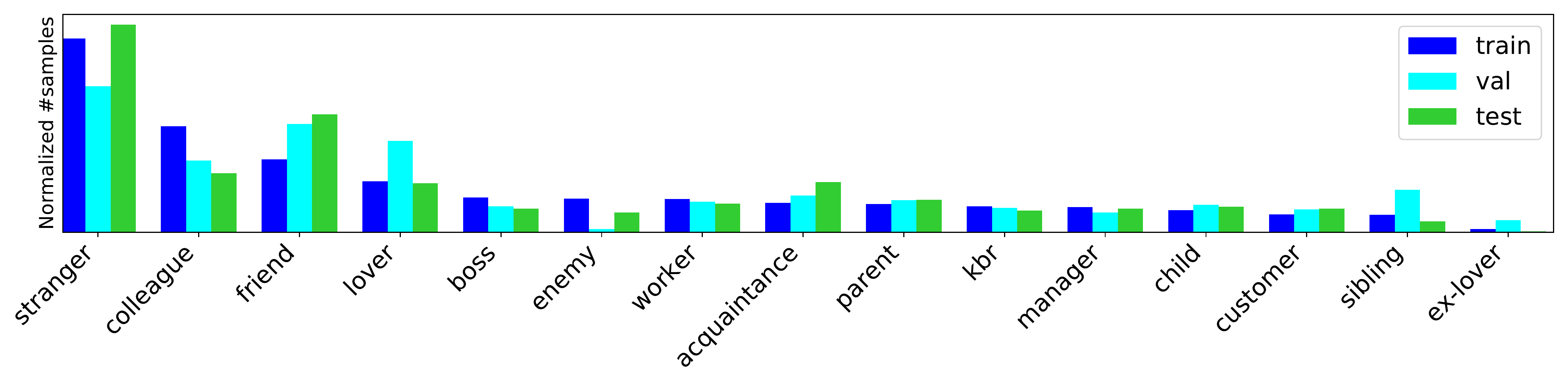}
\vspace{-0.5cm}
\caption{Distribution of relationship labels in train/val/test sets.
Sorted by descending order based on train set.}
\label{fig:rels_distr}
\vspace{4cm}
\end{figure*}

%%%%%%%%%%%%%%%%%%%%%%%%%%%%%%%%%%%%%%%%%%%%%%%%%%%%%%%%%%%%%%%%%%
% RADIAL TREES SHOWING GROUPING (INTERACTIONS, RELATIONSHIPS)
%%%%%%%%%%%%%%%%%%%%%%%%%%%%%%%%%%%%%%%%%%%%%%%%%%%%%%%%%%%%%%%%%%
\begin{figure*}
\centering
\includegraphics[width=\linewidth]{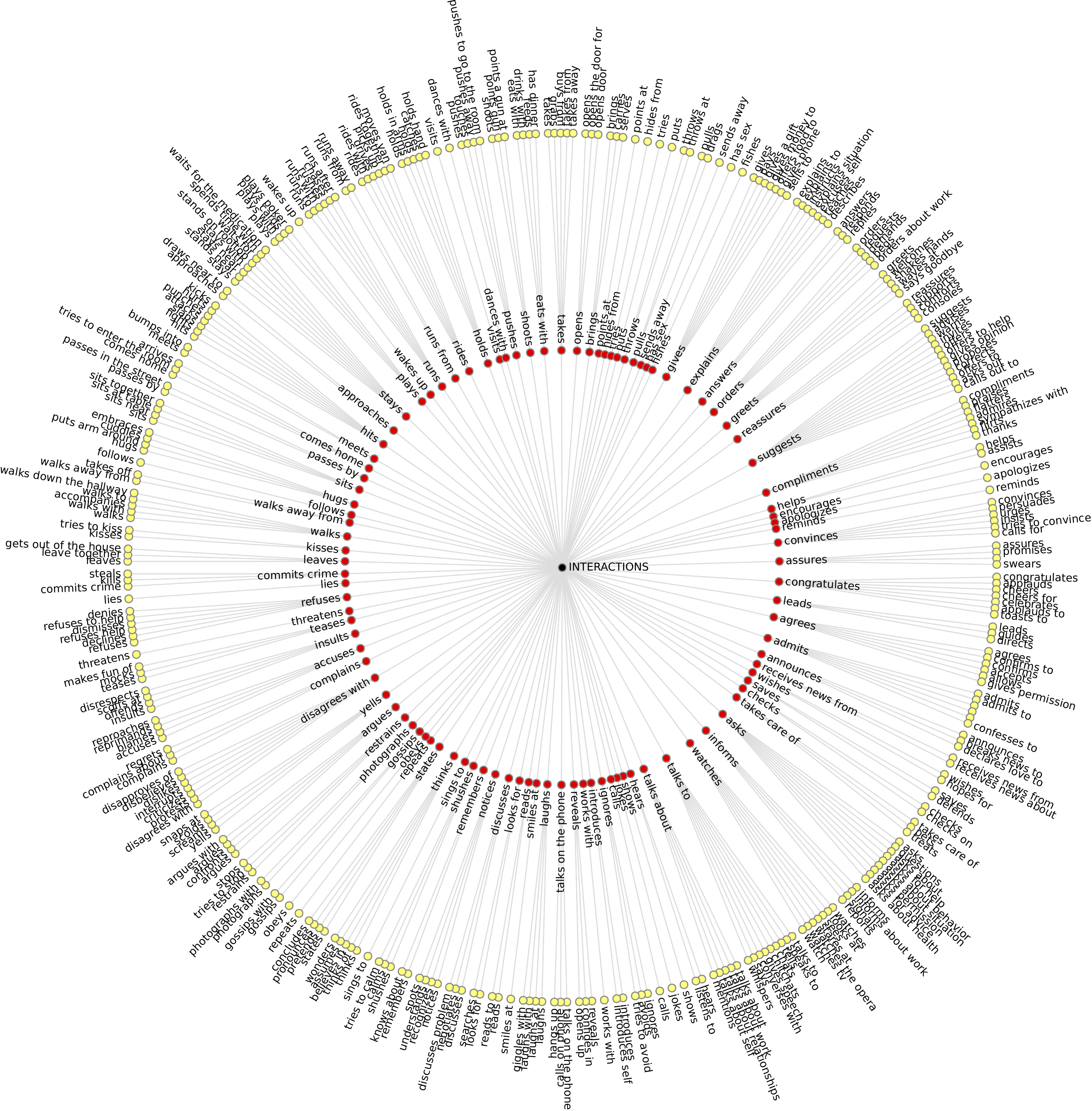}
\vspace{-0.5cm}
\caption{Diagram depicting how we group 324 interaction classes (outer circle) into 101 (inner circle). Best seen on the screen.}
\label{fig:ints_101}
\vspace{-0.4cm}
\end{figure*}

\begin{figure*}
\centering
\includegraphics[width=0.8\linewidth]{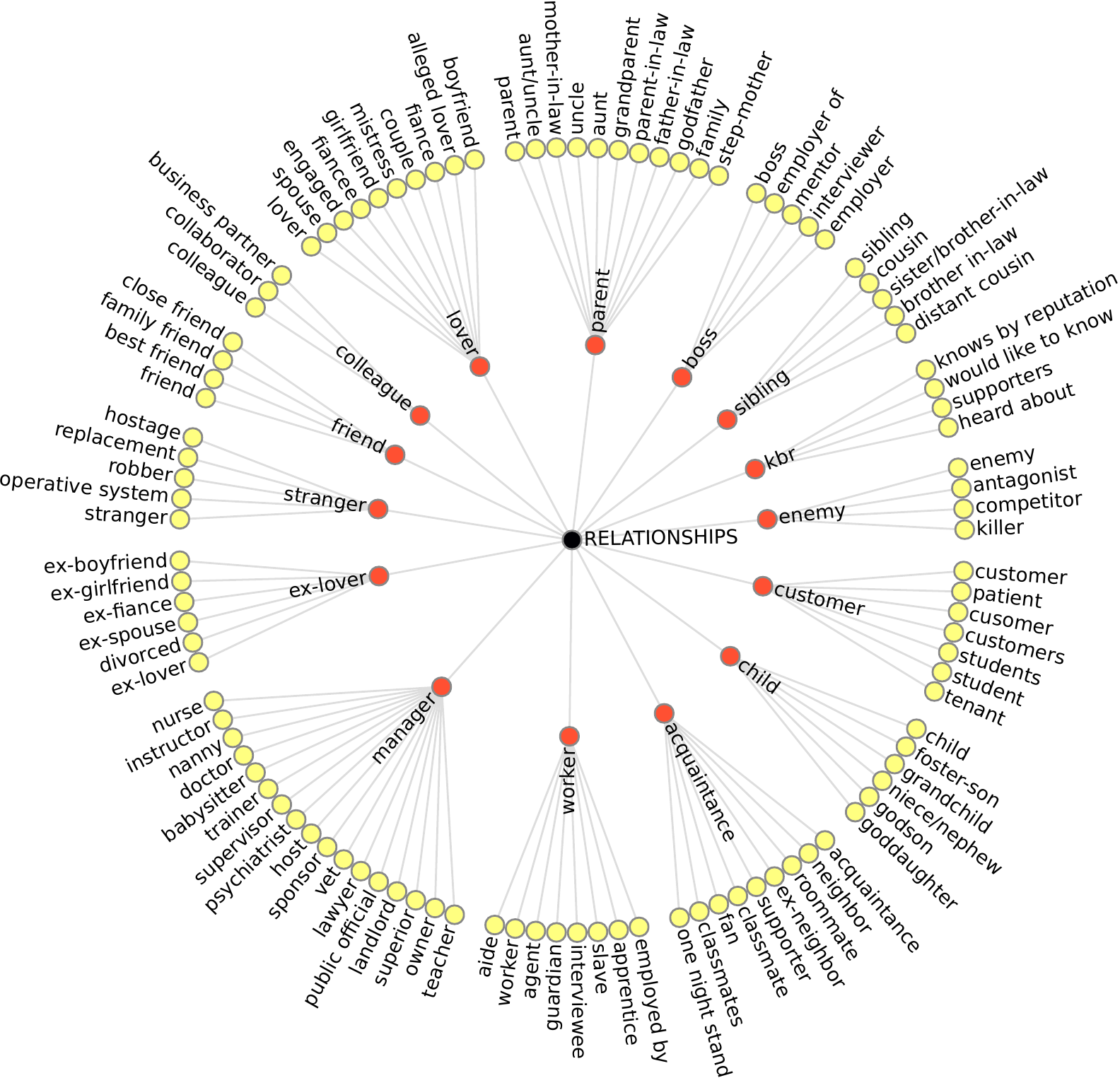}
\caption{Diagram depicting how we group 107 relationship classes (outer circle) into 15 (inner circle).}
\label{fig:rels_15}
\vspace{-0.4cm}
\end{figure*}

%%%%%%%%%%%%%%%%%%%%%%%%%%%%%%%%%%%%%%%%%%%%%%%%%%%%%%%%%%%%%%%%%%
% TEXT
%%%%%%%%%%%%%%%%%%%%%%%%%%%%%%%%%%%%%%%%%%%%%%%%%%%%%%%%%%%%%%%%%%

\section{Impact of Modalities}
We analyze the impact of modalities by presenting qualitative examples where using multiple modalities help predict the correct interactions.
Qualitative results presented here, refer to the quantitative performance indicated in Table~\ref{table:intr_only}.
Fig.~\ref{fig:v_m} shows that using dialog can help to improve predictions,
Fig.~\ref{fig:t_m} demonstrates the necessity of visual clip information and highlights that the two modalities are complementary.
Finally, Fig.~\ref{fig:m_tr} shows that focusing on tracks (visual representations in which the two characters appear) provides further improvements to our model.
Furthermore, Fig.~\ref{fig:top5_modals} shows top-5 interaction classes that benefit most from using additional modalities.

\vspace{1mm}
\noindent \textbf{Analyzing modalities.}
We also analyze the two models trained on only visual or only dialog cues (first two rows of Table~\ref{table:intr_only}).
Some interactions can be recognized only with \textbf{visual} (v) features:
\textit{rides} 63\% (v) / 0\% (d),
\textit{walks} 29\% (v) / 0\% (d),
\textit{runs} 26\% (v) / 0\% (d);
while others only with \textbf{dialog} (d) cues:
\textit{apologizes} 0\% (v) / 66\% (d),
\textit{compliments} 0\% (v) / 26\% (d),
\textit{agrees} 0\% (v) / 25\% (d).

Interactions that achieve non-zero accuracy with both modalities are:
\textit{hits} 64\% (v) / 5\% (d),
\textit{greets} 12\% (v) / 57\% (d),
\textit{explains} 25\% (v) / 51\% (d).

Additionally, the top-5 predicted classes for
\textbf{visual} cues are \emph{asks} 77\%, \emph{hits} 64\%, \emph{rides} 63\%, \emph{watches} 49\%, \emph{talks on phone} 41\%; and
\textbf{dialog} cues are \emph{asks} 75\%, \emph{apologizes} 66\%, \emph{greets} 57\%, \emph{explains} 51\%, \emph{watches} 30\%.
As \emph{asks} is the most common class, and \emph{watches} is the second most common, these interactions work well with both modalities.
% \emph{asks} works well for both models. % (\emph{watches} is second most common).

% classes which improves the most from modality vision to all together ...
% \begin{itemize}
%     \item \underline{5 top from visual to visual + textual}
%     \item \underline{5 top from visual+textual to with tracks}
% \end{itemize}

% + explain interactions more on these examples, talk a little about textual/visual classes

\section{Joint Interaction and Relationships}

\noindent \textbf{Confusion matrices.}
Fig.~\ref{fig:ints_conf_mat} shows the confusion matrix in the top-15 most commonly occurring interactions on the validation and test sets.
We see that multiple dialog based interactions (\eg~talks to, informs, and explains) are often confused.
We also present confusion matrices for relationships in Fig.~\ref{fig:rels_conf_mat}.
A large part of the confusion is due to lack of sufficient data to model the tail of relationship classes.

\vspace{1mm}
\noindent \textbf{Qualitative examples.}
Related to Table~\ref{table:joint_intr_reln_multiclip}, Fig.~\ref{fig:intr_mods} shows some examples where interaction predictions improve by jointly learning to model both interactions and relationships.
Similarly, Fig.~\ref{fig:rels_mods} shows how relationship classification benefits from our multi-task training setup.

% \begin{itemize}
%     \item \underline{confusion mat. interactions (val/test sets)}
%     \item \underline{confusion mat. relationships (val/test sets)}
%     \item \underline{interaction improves, 2 examples}
%     \item \underline{relationships improves, 2 examples}
%     makes sense to notice that relationships starts when two people strangers, so it means that there are quite a bit of examples where strangers are also kissing
% \end{itemize}

\section{Examples for \emph{Who} is Interacting}
Empirical evaluation shows that the knowledge about the relationship is important for localizing the pair of characters (Table~\ref{table:intr_tracks_reln}).
% 6 of the main paper).
In Fig.~\ref{fig:weak_tracks}, we illustrate an example where the dad walks into a room, sees his daughter \emph{with} someone, and asks questions (see figure caption for details).

Finally, in Fig.~\ref{fig:weak_tracks_joint}, we show an example where the model is able to correctly predict all components (interaction class, relationship type and the pair of tracks) in a complex situation with more than 2 people appearing in the clip.

% To show how track prediction improves when the model learns to identify along with \textit{who} (tracks) performing the interaction activity \textit{doing what?} (interaction class)
% \begin{itemize}
%     \item how improves track prediction when you use additionally gt relationship information, 2 examples
%     \item 1 example of joint prediction (mention that as qualitative result there is more than 2 persons) + that we consider pairs, and also A none, none A -> the task is more difficult if to consider only two persons which are with in the clip (due to missing annotations)
% \end{itemize}

\section{Dataset Analysis}
Fig.~\ref{fig:ints_distr} and Fig.~\ref{fig:rels_distr} show normalized distributions for the number of samples in each class for train, validation and test sets of interactions and relationships respectively.
As can be seen the most common classes appear many more times than the others.
% than the long tail of rare instances for all the sets.
Data from a complete movie belongs to one of the three train/val/test sets to avoid model bias on the plot and characters behaviour.
Notably, this means that the relative ratios between number of samples per class are also not necessarily consistent making the dataset and task even more challenging.

In the main paper, we described our approach to group over 300 interactions into 101 classes, and over 100 relationships into 15.
We use radial tree diagrams to depict the groupings for interaction and relationship labels, visualized in Fig.~\ref{fig:ints_101} and~\ref{fig:rels_15} respectively.